\documentclass{article}
\usepackage{conference,times}
\usepackage{hyperref}
\usepackage{url}
\usepackage{graphicx}
\usepackage{natbib}
\usepackage{caption}
\usepackage{algorithm}
\usepackage{algorithmic}
\usepackage{amsmath}
\usepackage{mathtools}
\usepackage{newfloat}
\usepackage{listings}
\usepackage{multirow}
\usepackage{booktabs}
\usepackage{tabularx}
\usepackage{adjustbox}
\usepackage{etoolbox}
\hypersetup{hidelinks}
\usepackage[table]{xcolor}
\usepackage[most]{tcolorbox}
\definecolor{PromptBg}{RGB}{248,250,252}
\definecolor{PromptFrame}{RGB}{80,110,150}
\definecolor{PromptBlue}{RGB}{30,90,170}
\definecolor{PromptGreen}{RGB}{20,120,80}
\definecolor{PromptRed}{RGB}{180,60,60}
\definecolor{casepurple}{HTML}{6B3FA0}
\definecolor{casepurplelight}{HTML}{F7F2FB}
\definecolor{caseblue}{HTML}{4C76D0}
\definecolor{casegreen}{HTML}{6FA553}
\definecolor{caseorange}{HTML}{E59A1F}
\definecolor{casegray}{HTML}{666666}
\definecolor{decisionred}{RGB}{180,40,40}
\newcommand{\skillfield}[1]{
\textbf{\scriptsize #1}:~
}
\newcommand{\skillcode}[1]{
{\ttfamily\scriptsize #1}
}
\newcommand{\skillstep}[1]{
\fbox{\ttfamily\scriptsize #1}
}
\newcommand{\field}[1]{
\textbf{\scriptsize #1}:~
}
\newcommand{\code}[1]{
{\ttfamily\scriptsize #1}
}
\lstdefinestyle{promptstyle}{
  basicstyle=\ttfamily\scriptsize,
  backgroundcolor=\color{PromptBg},
  frame=none,
  breaklines=true,
  breakatwhitespace=false,
  columns=fullflexible,
  keepspaces=true,
  showstringspaces=false,
  tabsize=2,
  xleftmargin=0pt,
  xrightmargin=0pt,
  aboveskip=0pt,
  belowskip=0pt
}
\tcbset{
promptbox/.style={
    enhanced,
    colback=PromptBg,
    colframe=PromptFrame,
    boxrule=0.5pt,
    arc=1mm,
    boxsep=0pt,
    left=1.5mm,
    right=1.5mm,
    top=1mm,
    bottom=1mm,
    before skip=0pt,
    after skip=0pt,
    fontupper=\scriptsize,
    before upper app={
        \setlength{\parindent}{0pt}
        \setlength{\parskip}{0pt}
    }
}
}
\tcbset{
figurebox/.style={
    enhanced,
    arc=1.5mm,
    boxrule=0.8pt,
    boxsep=0pt,
    left=1mm,
    right=1mm,
    top=0.8mm,
    bottom=0.8mm,
    fonttitle=\bfseries\scriptsize,
    coltitle=white,
    before upper app={
        \setlength{\parindent}{0pt}
    }
}
}
\tcbset{
skillcard/.style={
    enhanced,
    arc=1.5mm,
    boxrule=0.5pt,
    boxsep=0pt,
    left=0.8mm,
    right=0.8mm,
    top=0.3mm,
    bottom=0.3mm,
    before skip=0pt,
    after skip=0pt,
    fonttitle=\ttfamily\scriptsize,
    fontupper=\scriptsize,
    before upper app={
        \raggedright
        \setlength{\parindent}{0pt}
        \setlength{\parskip}{0pt}
    },
    valign=top
}
}
\tcbset{
atomcard/.style={
    enhanced,
    arc=1.5mm,
    boxrule=0.5pt,
    boxsep=0pt,
    left=0.8mm,
    right=0.8mm,
    top=0.3mm,
    bottom=0.3mm,
    before skip=0pt,
    after skip=0pt,
    fonttitle=\ttfamily\scriptsize,
    fontupper=\scriptsize,
    before upper app={
        \raggedright
        \setlength{\parindent}{0pt}
        \setlength{\parskip}{0pt}
    },
    valign=top
}
}
\tcbset{
casepanel/.style={
    enhanced,
    breakable,
    colback=white,
    colframe=casepurple!60,
    colbacktitle=casepurplelight,
    coltitle=casepurple,
    title filled=true,
    fonttitle=\bfseries\scriptsize,
    fontupper=\fontsize{7.5pt}{8.5pt}\selectfont,
    boxrule=0.5pt,
    arc=1mm,
    boxsep=0pt,
    left=1mm,
    right=1mm,
    top=0.5mm,
    bottom=0.5mm,
    before skip=0.3mm,
    after skip=0.3mm
}
}
\tcbset{
casestep/.style={
    enhanced,
    breakable,
    colback=white,
    colframe=casepurple!45,
    colbacktitle=casepurplelight,
    coltitle=casepurple,
    title filled=true,
    fonttitle=\bfseries\scriptsize,
    fontupper=\fontsize{7.5pt}{8.5pt}\selectfont,
    boxrule=0.4pt,
    arc=1mm,
    boxsep=0pt,
    left=1mm,
    right=1mm,
    top=0.5mm,
    bottom=0.5mm,
    before skip=0.3mm,
    after skip=0.3mm
}
}
\lstdefinestyle{casestate}{
  basicstyle=\ttfamily\fontsize{6.8pt}{7.5pt}\selectfont,
  breaklines=true,
  breakatwhitespace=false,
  columns=fullflexible,
  keepspaces=true,
  showstringspaces=false,
  frame=none,
  numbers=none,
  aboveskip=0pt,
  belowskip=0pt,
  moredelim=[is][\color{decisionred}\bfseries]{@@}{@@}
}
\newcommand{\casefield}[2]{
{\scriptsize\textbf{#1}}
&
{\scriptsize #2}
\\[-0.15em]
}

\title{HiSkill: Empowering LLM Agents with Hierarchical Skill Graphs}

\author{
Yu Hao$^{1}$,
Jinxuan Cai$^{1}$,
Qi Zhang$^{2}$,
Yawen Li$^{1}$,
Zhiqiang Zhang$^{3}$,
Chuan Shi$^{1}$,
Cheng Yang$^{1\dagger}$\\
$^{1}$Beijing University of Posts and Telecommunications\\
$^{2}$China Mobile Group Shaanxi Co., Ltd\\
$^{3}$Ant Group\\
\texttt{
haoyuu@bupt.edu.cn,
caijinxuan@bupt.edu.cn,
zhangqi3@sn.chinamobile.com
}\\
\texttt{
warmly0716@126.com,
lingyao.zzq@antfin.com,
shichuan@bupt.edu.cn,
}\\
\texttt{
yangcheng@bupt.edu.cn
}
}

\footnotetext[2]{Corresponding author.}

\makeatletter
\iclrfinalcopy

\makeatother

\begin{document}
\pagestyle{empty}
\thispagestyle{empty}
\setlength{\tabcolsep}{3pt}
\renewcommand{\arraystretch}{1.05}
\AtBeginEnvironment{table}{\scriptsize\setlength{\tabcolsep}{0.5pt}\renewcommand{\arraystretch}{0.78}}
\renewcommand{\arraystretch}{0.95}
\maketitle

\begin{abstract}
Skills have become an important abstraction for enabling large language model (LLM) agents to reuse past experience in long-horizon interactive tasks. However, existing trajectory-to-skill methods often produce flat collections of high-level textual skills that are stored and retrieved independently, leaving skill relations underutilized and maintaining a gap between high-level skills and executable actions. 
In this paper, we propose HiSkill, a hierarchical skill graph framework that organizes interaction  trajectories into a directed graph with skill nodes, AtomicOp nodes, and typed edges. Specifically, the graph connects reusable high-level skills with executable action templates, while also capturing decomposition, temporal transition, compatibility, support, and recovery relations among them. 
At inference time, HiSkill retrieves a compact task-relevant subgraph and performs subgraph-guided task execution, where a symbolic task state, an active skill, and the retrieved subgraph guide the LLM agent to switch skills, select AtomicOps, and ground executable actions iteratively. 
Experiments on three interactive environments show that HiSkill outperforms state-of-the-art baselines while reducing inference token consumption, demonstrating the effectiveness of bridging high-level skills and executable action grounding through a hierarchical skill graph.
\end{abstract}

\section{Introduction}

Long-horizon interactive tasks require large language model (LLM) agents to solve multi-step problems through continuous interaction, where reusing historical experience has become an important way to improve future decision making~\citep{zhao2024expel,chhikara2025mem0,fang2025memp}.

Skills have recently emerged as a key abstraction for organizing such reusable experience. A skill is commonly defined as an on-demand modular capability unit that provides procedural knowledge, such as description, instructions, or metadata to guide future decision making~\citep{xu2026agent,anthropic2025equipping,liang2026skillnet}. 
Compared with one-off prompts, skills allow LLM agents to avoid repeatedly solving similar tasks from scratch, thereby improving the efficiency and stability of multi-step task solving.

\begin{figure}[t]
  \centering
  \includegraphics[width=\linewidth]{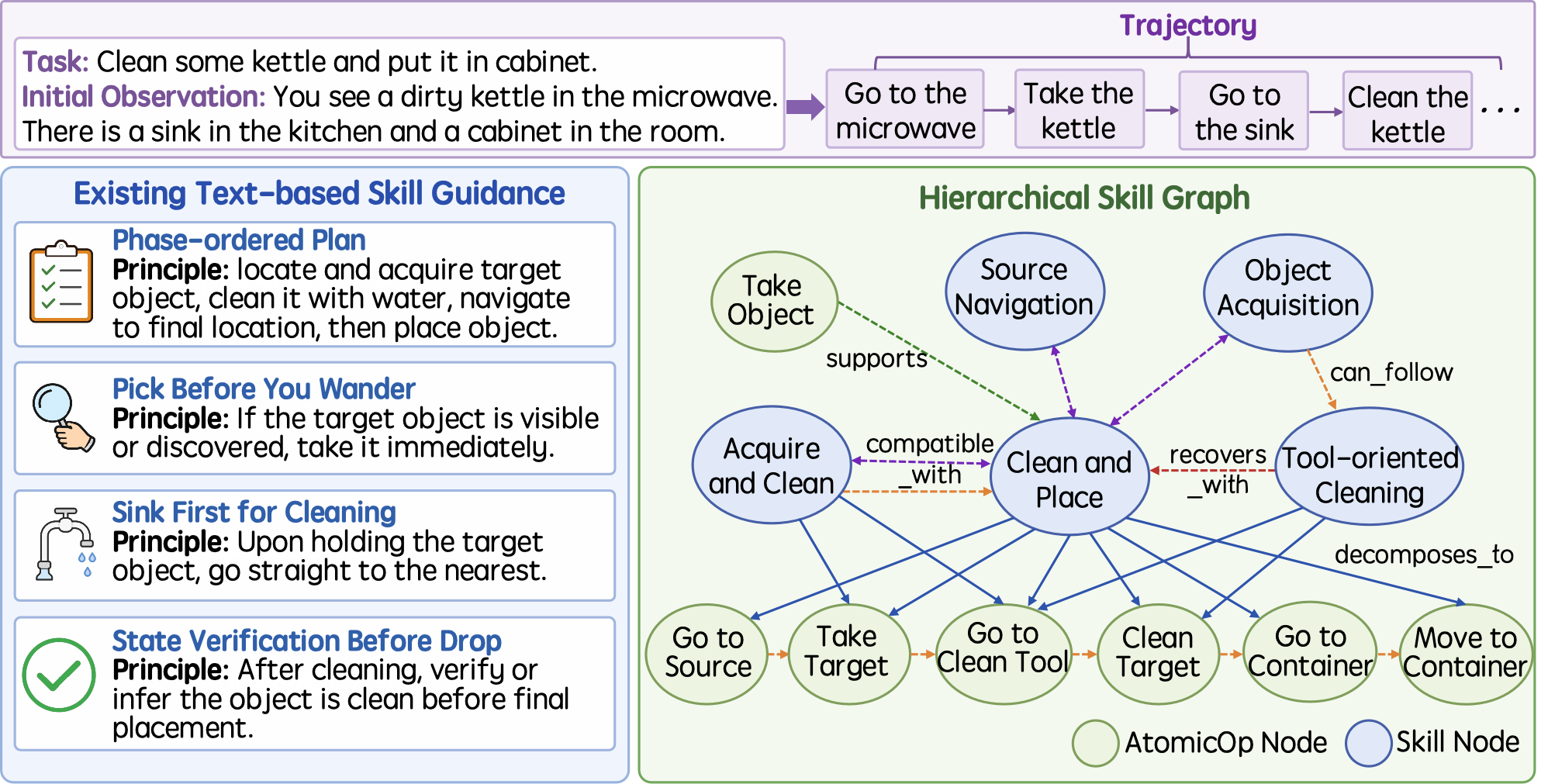}
  \caption{Existing methods distill trajectories into coarse-grained textual rules in a flat organization, while HiSkill builds a hierarchical graph to assist task execution, including high-level skill nodes, executable AtomicOp nodes and typed edges between them.}
  \label{fig:intro}
\end{figure}

To reduce the cost of skill construction, recent methods propose to automatically extract skills from interaction trajectories.
For example, SkillRL distills noisy trajectories into a SkillBank and recursively evolves it through reinforcement learning~\citep{xia2026skillrl}.
D2Skill organizes reusable experience into skills for task guidance and error correction~\citep{tu2026dynamic}. 
CoEvoSkills automatically generates skill packages and iteratively refines them through a skill generator and a surrogate verifier~\citep{zhang2026coevoskills}. 
The resulting skills are typically organized as flat collections of high-level textual descriptions.

However, existing trajectory-to-skill methods have two limitations. 
First, most generated skills remain coarse-grained strategies or textual guidance, leaving a gap between high-level skill descriptions and concrete executable actions. 
A skill may describe what subgoal to achieve or what general procedure to follow, but actual interaction requires deciding the next concrete action, with arguments and execution timing.  
Simply injecting high-level skill text into the LLM context does not always allow the model to infer these details~\citep{liu2026harnessing,li2026skillsbench}.
Second, many procedural steps are shared across tasks, but existing methods often construct and retrieve skills as independent units, underutilizing relations among skills and shared action patterns.

Recent graph-based methods highlight the importance of structure for skill use. For example, 
Graph of Skills retrieves dependency-aware skill bundles from an offline skill graph and shows that flat vector retrieval can miss execution dependencies~\citep{liu2026graph}. 
GraSP compiles retrieved skills into typed directed acyclic graphs and emphasizes that the bottleneck has shifted from skill availability to skill orchestration~\citep{xia2026grasp}. 
Nevertheless, these methods typically construct graphs at the skill level, and fail to explore the diverse ways in which skills may relate during execution. As a result, relations among skills are not fully exploited, and the retrieved skills are mainly high-level textual guidance, leaving the connection to executable actions implicit.

To address these limitations, we propose \textbf{HiSkill}, a method that constructs hierarchical skill graphs from trajectories and uses them for interactive task solving. 
As shown in Figure~\ref{fig:intro}, existing methods produce coarse-grained textual skills as high-level guidance, while HiSkill represents the trajectories as a hierarchical graph with skill nodes, AtomicOp nodes, and typed edges. 
Specifically, AtomicOp nodes denote executable action templates, while skill nodes denote reusable procedures composed of multiple AtomicOp nodes and enriched with structured metadata. 
The typed edges further capture procedural relations, allowing the graph to encode not only which procedures usually work, but also which skills can be composed, which auxiliary actions are needed, and which operations can be tried when execution fails.

At inference time, HiSkill retrieves a compact task-relevant subgraph through semantic--lexical matching and graph expansion. 
It then monitors a symbolic task state and an active skill, and uses them together with the retrieved subgraph and an LLM to switch or maintain the active skill, select an AtomicOp, or directly generate an action by an LLM. 
The resulting action is executed, and the new observation iteratively updates the task state for subsequent skill switching and AtomicOp selection. 
This design effectively bridges high-level skill abstraction and executable action grounding.

We summarize our main contributions as follows:
\begin{itemize}
\item We propose a hierarchical skill graph framework that converts trajectories into a directed graph with skill nodes, AtomicOp nodes and typed edges, bridging high-level skill abstraction and executable action templates.

\item We design a pipeline that constructs hierarchical skill graphs, retrieves task-relevant subgraphs, and performs subgraph-guided task execution.

\item We compare HiSkill with state-of-the-art baselines across three interactive environments. HiSkill achieves average relative improvements of 17.33\% in success rate, while reducing inference token consumption by 78.75\% over the strongest baseline.
\end{itemize}

\section{Related Work}

\subsection{Agentic Skills}
\label{sec:related-skills}

Agentic skills are reusable procedural capability units that package task-specific description, instructions, or metadata for LLM agents~\citep{xu2026agent,anthropic2025equipping,lu2026skill0}. Early and deployed skill systems often rely on manual authoring, such as Anthropic Agent Skills organized through \texttt{SKILL.md} files~\citep{anthropic2025agentskillsdocs} and OpenAI Custom GPTs / GPT Actions built from custom instructions, knowledge files, and API schemas~\citep{openai2024gptactions}. 
Manual skills are explicit and controllable, but scaling them to complex interactive environments requires substantial effort to specify procedures, applicability conditions, and execution details.
To reduce this manual cost, recent methods automatically construct skills from trajectories and interaction experience. 
Voyager maintains an ever-growing executable code skill library in Minecraft~\citep{wang2024voyager}; SkillRL and D2Skill distill trajectories into skill banks with different forms~\citep{xia2026skillrl,tu2026dynamic}; SkillX and CoEvoSkills study skill knowledge-base construction and automatic skill package generation~\citep{wang2026skillx,zhang2026coevoskills}; and CUA-Skill represents GUI operation knowledge as parameterized skills for computer-use agents~\citep{chen2026cua}.

Beyond individual skill construction, recent work also studies skill evaluation, ecosystem organization, and structured use. SkillsBench evaluates the effectiveness of curated and self-generated skills~\citep{li2026skillsbench}; SkillNet and AgentSkillOS organize skills as large-scale assets with ontology, evaluation, and orchestration mechanisms~\citep{liang2026skillnet,li2026organizing}; and Graph of Skills and GraSP introduce graph-structured retrieval or composition for large skill libraries~\citep{liu2026graph,xia2026grasp}. 
In addition, platforms such as SkillsMP, ClawHub, and LangSkills support skill packaging, publishing, browsing, and searching at scale, reflecting the shift from individual skills to ecosystem-level skill management~\citep{skillsmp2026,openclaw2026,labrai2026langskills}.
However, skills often remain at the level of coarse-grained strategies or textual guidance in existing systems. This leaves a gap between high-level skill descriptions and executable actions. Besides, the diverse ways in which skills may relate during execution are also underexplored.

\begin{figure*}[t]
  \centering
  \setlength{\tabcolsep}{6pt}
  \begin{tabular}{@{}c@{}}
    \includegraphics[width=1.0\linewidth]{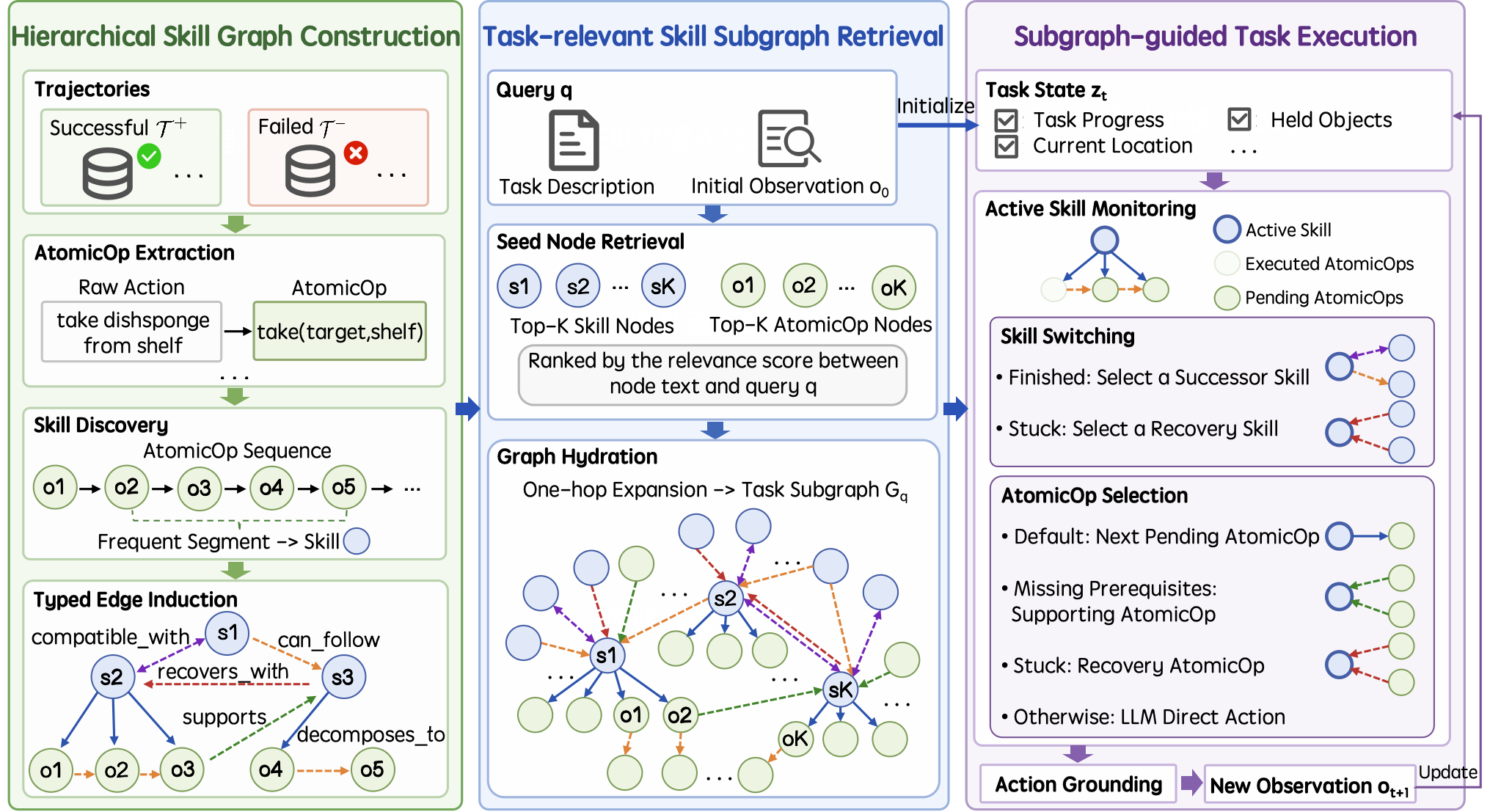}  
    \\[0.4em]
  \end{tabular}
  \caption{The overall framework of our proposed HiSkill.} 
  \label{fig:method} 
\end{figure*}

\subsection{Experience and Memory for LLM Agents}
\label{sec:related-memory}
Experience and memory mechanisms allow LLM agents to reuse past interactions. For example, Reflexion stores verbal feedback for self-improvement~\citep{shinn2023reflexion}, ExpeL extracts natural-language insights from past tasks~\citep{zhao2024expel}, and Mem0 studies scalable long-term memory extraction and retrieval for agents~\citep{chhikara2025mem0}. 
More closely related to task execution, MemP explores procedural memory by distilling past trajectories into step-level instructions and script-like abstractions~\citep{fang2025memp}.

Overall, memory-based methods demonstrate the value of reusing historical experience, but they typically store experience as reflections, text memories, summaries, or procedural instructions. They pay less attention to organizing high-level procedures, executable actions, and execution relations into a unified structure for runtime action grounding.

\section{Methodology}
\label{sec:method}

\subsection{Framework Overview}
\label{sec:overview}

As shown in Figure~\ref{fig:method}, HiSkill consists of three main components:
(1) \emph{Hierarchical Skill Graph Construction} builds a directed graph from interaction trajectories, where AtomicOp nodes denote executable action templates, skill nodes denote high-level procedures composed of multiple AtomicOp nodes and structured metadata, and typed edges capture relations among them. 
(2) \emph{Task-relevant Skill Subgraph Retrieval} retrieves a compact task-relevant subgraph from the constructed skill graph according to semantic similarity, lexical matching, and graph relations. 
(3) \emph{Subgraph-guided Task Execution} monitors the current task state and active skill, and uses the retrieved subgraph as well as an LLM to switch skills or select AtomicOps for task execution.

\subsection{Hierarchical Skill Graph Construction}
\label{sec:skill-graph-construction}

We first construct a hierarchical skill graph from historical interaction trajectories. 
Formally, we partition the trajectory set $\mathcal{T}$ into:
\begin{equation}
\mathcal{T}^{+}=\{\tau \in \mathcal{T}: r(\tau)=1\}, 
\quad 
\mathcal{T}^{-}=\{\tau \in \mathcal{T}: r(\tau)=0\},
\label{eq:traj-split}
\end{equation}
where $r(\tau)\in\{0,1\}$ indicates whether $\tau$ successfully completes the task. Successful trajectories  $\mathcal{T}^{+}$ provide reusable execution patterns, while failed trajectories $\mathcal{T}^{-}$ provide failure modes and recovery evidence. 
Now we will briefly introduce how to extract atomic operations, discover skills, and induce typed edges from these trajectories. Implementation details are provided in Appendix~\ref{app:graph-construction-details}.

\subsubsection{AtomicOp Extraction.}
We extract raw action sequences from interaction trajectories and canonicalize each raw action into an AtomicOp occurrence. 
Specifically, a schema-based canonicalizer parses the raw action, identifies its action type and arguments, replaces task-specific entities with placeholders, and produces a normalized action template. 
We then deduplicate and merge AtomicOp occurrences with the same normalized template, while accumulating their support counts across trajectories. 
The resulting AtomicOp node set $V_O$ serves as the executable action vocabulary for later skill discovery and action grounding.

\subsubsection{Skill Discovery.}
We then mine skills from the AtomicOp sequences of successful trajectories. 
Specifically, we identify frequent action patterns, and abstract each retained pattern into a skill node, forming the skill set $V_S$. 
Each skill stores an ordered AtomicOp sequence and structured metadata, including applicability states, expected state changes, representative examples, argument candidates, support count, and failure hints. 
Therefore, a skill is not merely a textual hint, but a reusable procedural abstraction grounded in observed action patterns. 
To improve readability, we use an LLM only to refine natural-language fields such as names and descriptions, without changing the mined AtomicOp structure. 

\paragraph{Typed Edge Induction.}
Finally, we induce typed edges from successful and failed trajectories to construct a hierarchical skill graph:
\begin{equation}
G=(V,E), \quad V=V_S \cup V_O,\quad E\subseteq V\times V\times \mathcal{L},
\label{eq:skill-graph}
\end{equation}
where $V_S$ and $V_O$ denote the sets of skill and AtomicOp nodes, and each edge $(u,v,\ell)\in E$ denotes a directed relation of type $\ell$ from node $u$ to node $v$. 
The relation type set is:
\begin{equation}
\begin{aligned}
\mathcal{L}=\{&
\mathtt{decomposes\_to},\ 
\mathtt{can\_follow},\
\mathtt{compatible\_with},\
\mathtt{supports},\
\mathtt{recovers\_with}
\}.
\end{aligned}
\label{eq:edge-labels}
\end{equation}

These edges are induced from trajectory statistics, state compatibility, failure patterns, and semantic matching. 
Specifically, $\mathtt{decomposes\_to}$ edges link each skill node to the ordered AtomicOp nodes, specifying how a skill is decomposed into executable action templates;  
$\mathtt{can\_follow}$ edges connect skill-to-skill or AtomicOp-to-AtomicOp nodes, and are induced by counting ordered co-occurrences in successful skill or AtomicOp sequences, capturing directed temporal continuations;  
$\mathtt{compatible\_with}$ edges connect skill nodes that frequently co-occur in successful trajectories and have compatible pre/post-states, capturing unordered skill composition as symmetric directed edges; 
$\mathtt{supports}$ edges link auxiliary AtomicOp nodes to skill nodes when the AtomicOp frequently appears in the context of a skill and helps satisfy missing prerequisites or provides supporting state evidence; 
$\mathtt{recovers\_with}$ edges connect recovery skill/AtomicOp nodes to failed or stalled skill/AtomicOp nodes, and are inferred from failed trajectories, repeated ineffective actions, failure hints, missing-state evidence, and semantic/state matching.

In summary, successful trajectories mainly induce decomposition, transition, compatibility, and support relations, while failed trajectories mainly induce recovery relations. 
The resulting hierarchical skill graph encodes high-level skills, executable AtomicOps, and their relations, providing a structured foundation to assist task execution.

\subsection{Task-relevant Skill Subgraph Retrieval}
\label{sec:subgraph-retrieval}

After constructing the hierarchical skill graph $G=(V,E)$, we retrieve a compact task-relevant subgraph for each task $q$ instead of using the full graph:
\begin{equation}
G_q=(V_q,E_q),\quad V_q \subseteq V,\quad E_q \subseteq E.
\label{eq:retrieved-subgraph}
\end{equation}
The retrieval process first selects task-relevant skill and AtomicOp seed nodes, and then enriches them with graph relations needed for execution.
Implementation details are provided in Appendix~\ref{app:subgraph-retrieval-details}.

\subsubsection{Seed Node Retrieval.}
We construct a retrieval query $x_q$ by concatenating the task description and initial observation. 
For each node $v \in V$, we compute a relevance score using its textualized representation $d_v$.
Following the hybrid retrieval strategy used in ToolScope~\citep{liu2026toolscope} and GoS~\citep{liu2026graph}, we combine dense semantic similarity and sparse lexical matching:
\begin{equation}
\mathrm{score}(v \mid q)
=
\lambda \, \mathrm{Dense}(x_q,d_v)
+
(1-\lambda)\,\mathrm{Sparse}(x_q,d_v),
\label{eq:node-retrieval-score}
\end{equation}
where $\mathrm{Dense}(\cdot)$ computes cosine similarity between query and node embeddings, $\mathrm{Sparse}(\cdot)$ is  BM25~\citep{robertson2009probabilistic} score based on lexical term overlap, and $\lambda\in[0,1]$ balances the two signals.  
We rank skill and AtomicOp nodes separately according to Eq.~\eqref{eq:node-retrieval-score}, and retain the top-$K$ nodes from each type, and denote their union as $V_{\mathrm{seed}}$.

\subsubsection{Graph Hydration.}
After obtaining the seed nodes, we enrich them with one-hop expansion to form the task-relevant subgraph. 
Considering the directionality of different edge types, the expanded node set is defined as:
\begin{equation}
\begin{aligned}
V_q
=
&V_{\mathrm{seed}} \
\cup
\{v \mid (u,v,\ell)\in E,\ u\in V_{\mathrm{seed}},
\ell=\mathtt{decomposes\_to}\}\\
&\cup
\{v \mid (v,u,\ell)\in E,\ u\in V_{\mathrm{seed}},
\ell\neq\mathtt{decomposes\_to}\}.
\end{aligned}
\label{eq:subgraph-expansion}
\end{equation}

Then we retain all typed edges among the selected nodes 
$E_q=\{(u,v,\ell)\in E \mid u\in V_q, v\in V_q\}$. The resulting subgraph contains task-relevant skill nodes, AtomicOp nodes, and their typed relations for runtime execution.

\subsection{Subgraph-guided Task Execution}
\label{sec:subgraph-task-execution}

The retrieved subgraph provides candidate skill and AtomicOp nodes for assisting task execution. At runtime, the system performs stepwise task execution by monitoring the current task state and active skill: the task state records the current interaction status, and the active skill serves as the current execution anchor. 
At each step, the system updates the state, switches or keeps the active skill, selects a relevant AtomicOp target, and grounds it into a concrete executable action. 
Implementation details are provided in Appendix~\ref{app:subgraph-execution-details}.

\subsubsection{Task State.}

At step $t$, the system maintains a task state $z_t$, implemented as a symbolic key-value memory. 
After executing a concrete action $a_t$ and receiving the next observation $o_{t+1}$, the state is updated as:
\begin{equation}
z_{t+1}=\mathrm{UpdateState}(z_t,a_t,o_{t+1}).
\label{eq:state-update}
\end{equation}

The state records execution-relevant evidence, such as task progress, held objects, current location, and visited entities. 
It is later used to check skill completion and applicability, detect missing prerequisites, and identify stagnation or repeated failures.

\subsubsection{Skill Switching.}

At step $t$, the system maintains an active skill node $s_t\in V_q\cap V_S$. 
The active skill node is associated with an ordered sequence of AtomicOp nodes through its $\mathtt{decomposes\_to}$ edges. 
The initial skill node is selected from $V_q$ by the LLM according to the task, initial observation, and task state. The active skill will be switched in two cases: when the active skill is finished, a successor skill node is selected from $V_q$ through $\mathtt{can\_follow}$ or $\mathtt{compatible\_with}$ edges; 
when execution is stuck or repeatedly fails, the system switches to a recovery skill node through $\mathtt{recovers\_with}$ edges.

\subsubsection{AtomicOp Selection.}

By default, it selects the next unexecuted AtomicOp node from the active skill through $\mathtt{decomposes\_to}$ edges; 
when required prerequisites are missing, it selects a support AtomicOp node through $\mathtt{supports}$ edges; 
when execution is stuck or repeatedly fails, it selects a recovery AtomicOp node through $\mathtt{recovers\_with}$ edges; 
if no subgraph-related action is suitable, the system allows an LLM direct-action option. 
We denote the selected operation target as $u_t$, which can be either an AtomicOp node or an LLM-generated action.

\subsubsection{Action Grounding and State Update.}

Given the selected operation target $u_t$, the system grounds it into a concrete executable action according to the current observation and task state:
\begin{equation}
a_t
=
\mathrm{Grounding}(u_t,o_t,z_t),
\label{eq:action-grounding}
\end{equation}
where $a_t$ denotes the executable action with instantiated arguments.

After executing $a_t$, the system receives a new observation $o_{t+1}$ and updates the task state according to Eq.~\eqref{eq:state-update}. 
The updated state is then used for subsequent skill switching and AtomicOp selection. This iterative process continues until the task is completed.

\begin{table}[t]
\centering
\scriptsize
\setlength{\tabcolsep}{1.0pt}
\renewcommand{\arraystretch}{1.08}

\resizebox{\linewidth}{!}{
\begin{tabular}{lccccccc|ccccccc}
\toprule
\multirow{3}{*}{\textbf{Method}}
& \multicolumn{14}{c}{\textbf{ALFWorld}}\\
\cmidrule(lr){2-15}

& \multicolumn{7}{c}{\textbf{Seen}}
& \multicolumn{7}{c}{\textbf{Unseen}}\\

\cmidrule(lr){2-8}
\cmidrule(lr){9-15}

& \textbf{Pick} & \textbf{Look} & \textbf{Clean} & \textbf{Heat} & \textbf{Cool} & \textbf{Pick2} & \textbf{All}
& \textbf{Pick} & \textbf{Look} & \textbf{Clean} & \textbf{Heat} & \textbf{Cool} & \textbf{Pick2} & \textbf{All} \\
\midrule

\rowcolor{gray!16}
\multicolumn{15}{l}{\textit{\textbf{Prompt-based Methods}}}\\

ReAct
& 82.86 & 84.62 & 59.26 & 0.00 & 24.00 & 87.50 & 59.29
& 87.50 & \textbf{100.00} & \underline{93.55} & 8.70 & 57.14 & 64.71 & 66.42 \\
Reflexion
& 85.71 & \textbf{100.00} & 62.96 & 68.75 & 84.00 & 58.33 & 75.71
& \underline{95.83} & 83.33 & 83.87 & 43.48 & 90.48 & 82.35 & 79.85 \\

\midrule

\rowcolor{gray!16}
\multicolumn{15}{l}{\textit{\textbf{Memory-based Methods}}}\\

ExpeL
& 77.14 & \textbf{100.00} & 70.37 & 6.25 & 48.00 & 41.67 & 58.57
& 83.33 & 66.67 & 80.65 & 30.43 & 66.67 & 82.35 & 68.66 \\
Mem0
& \underline{94.29} & 92.31 & 77.78 & 37.50 & 76.00 & 87.50 & 80.00
& 91.67 & 94.44 & \underline{93.55} & 39.13 & 85.71 & \textbf{100.00} & 83.58 \\
MemP
& 88.57 & \textbf{100.00} & \underline{88.89} & \textbf{75.00} & \underline{92.00} & 83.33 & \underline{87.86}
& \underline{95.83} & \textbf{100.00} & \textbf{100.00} & \underline{60.87} & \underline{95.24} & 70.59 & \underline{88.06} \\
SimpleMem
& 86.11 & 86.67 & 62.96 & 0.00 & 20.00 & 82.61 & 60.71
& 92.00 & 83.33 & 83.87 & 4.55 & 33.33 & 76.47 & 63.43 \\

\midrule

\rowcolor{gray!16}
\multicolumn{15}{l}{\textit{\textbf{Skill-based Methods}}}\\

Vector Skills
& 85.71 & 76.92 & 66.67 & 6.25 & 52.00 & \underline{91.67} & 67.14
& \underline{95.83} & 88.89 & 70.97 & 4.35 & 47.62 & 64.71 & 61.94 \\
SkillNet
& 91.43 & 61.54 & 66.67 & 68.75 & 80.00 & 75.00 & 76.43
& \underline{95.83} & 83.33 & 83.87 & 47.83 & 76.19 & 76.47 & 77.61 \\
GoS
& 88.57 & 92.31 & 66.67 & 6.25 & 24.00 & 83.33 & 62.86
& \underline{95.83} & 94.44 & \underline{93.55} & 4.35 & 38.10 & 58.82 & 65.67 \\
\textbf{HiSkill}
& \textbf{100.00} & \textbf{100.00} & \textbf{92.59} & \textbf{75.00} & \textbf{96.00} & \textbf{95.83} & \textbf{94.29}
& \textbf{100.00} & \textbf{100.00} & 87.10 & \textbf{69.57} & \textbf{100.00} & \textbf{100.00} & \textbf{91.79} \\

\bottomrule
\end{tabular}
}

\caption{Performance on ALFWorld with Gemini-2.5-Pro: Success rate (\%) is reported. The best results are highlighted in boldface, and the second-best results are underlined.}
\label{tab:alfworld_seen_unseen}

\end{table}

\begin{table}[t]
\centering
\small
\setlength{\tabcolsep}{1.7mm}
\renewcommand{\arraystretch}{1.10}
\begin{tabular}{lcc|cc|cc}
\toprule
\multirow{3}{*}{\textbf{Method}}
& \multicolumn{2}{c|}{\textbf{WebShop}}
& \multicolumn{4}{c}{\textbf{ScienceWorld}} \\
\cmidrule(lr){2-3} \cmidrule(lr){4-7}
& \multirow{2}{*}{\textbf{Score}}
& \multirow{2}{*}{\textbf{Succ.}}
& \multicolumn{2}{c|}{\textbf{Seen}}
& \multicolumn{2}{c}{\textbf{Unseen}} \\
\cmidrule(lr){4-5} \cmidrule(lr){6-7}
&
&
& \textbf{Score}
& \textbf{Succ.}
& \textbf{Score}
& \textbf{Succ.} \\
\midrule

\rowcolor{gray!16}
\multicolumn{7}{l}{\textit{\textbf{Prompt-based Methods}}} \\
ReAct
& 51.46 & 34.40
& 39.36 & 34.54
& 39.65 & 38.86 \\
Reflexion
& 35.88 & 29.20
& \underline{66.74} & 63.92
& 58.96 & 52.61 \\

\midrule
\rowcolor{gray!16}
\multicolumn{7}{l}{\textit{\textbf{Memory-based Methods}}} \\
ExpeL
& \underline{69.73} & \underline{55.40}
& 63.73 & 63.92
& 58.23 & 59.24 \\
Mem0
& 46.00 & 29.20
& 56.48 & 51.55
& 52.84 & 50.71 \\
MemP
& 53.33 & 39.20
& 62.91 & \underline{64.43}
& \underline{67.20} & \underline{61.14} \\
SimpleMem
& 50.13 & 32.00
& 64.05 & 61.86
& 57.95 & 52.13 \\

\midrule
\rowcolor{gray!16}
\multicolumn{7}{l}{\textit{\textbf{Skill-based Methods}}} \\
Vector Skills
& 51.84 & 36.00
& 30.13 & 26.80
& 35.25 & 27.49 \\
SkillNet
& 40.57 & 35.60
& 60.82 & 58.69
& 59.11 & 56.40 \\
GoS
& 52.15 & 36.80
& 36.81 & 38.14
& 39.72 & 41.71 \\
\textbf{HiSkill}
& \textbf{77.50} & \textbf{67.60}
& \textbf{88.72} & \textbf{84.02}
& \textbf{84.20} & \textbf{81.04} \\

\bottomrule
\end{tabular}
\caption{Performance on WebShop and ScienceWorld with Gemini-2.5-Pro: Average score and success rate (\%) are reported. The best results are highlighted in boldface, and the second-best results are underlined.}
\label{tab:webshop_scienceworld_results}
\end{table}

\section{Experiments}

We conduct extensive experiments to answer the following research questions (RQs):
\textbf{RQ1:} How does the proposed HiSkill perform compared with state-of-the-art baselines?
\textbf{RQ2:} How does HiSkill perform in terms of inference token consumption?
\textbf{RQ3:} How do the key components contribute to the overall effectiveness?
\textbf{RQ4:} How sensitive is HiSkill to key hyperparameters?
Besides, we provide a case study in Appendix~\ref{app:case-study}.

\subsection{Experimental Setup}

\subsubsection{Datasets.}
We evaluate our method on three interactive environments: ALFWorld~\citep{shridhar2020alfworld}, WebShop~\citep{yao2022webshop}, and ScienceWorld~\citep{wang2022scienceworld}. These environments cover embodied household manipulation, web shopping, and scientific experimentation, respectively, and require multi-step decision making under textual observations.
Detailed dataset descriptions and statistics are provided in Appendix~\ref{app:dataset-details}.

\subsubsection{Evaluation.}
For ALFWorld and WebShop, we follow the evaluation settings used in GiGPO~\citep{feng2025group} and SkillRL~\citep{xia2026skillrl}. For ScienceWorld, we follow ETO~\citep{song2024trial} and SkillNet~\citep{liang2026skillnet}. 
ALFWorld and ScienceWorld are evaluated on both Seen and Unseen splits. 
We further hold out 10\% of the training set as a validation split for hyperparameter tuning and use the remaining training data for skill graph construction. 
Since ALFWorld provides a binary task completion signal, we report success rate (\%) for each subtask and the overall average. 
For WebShop and ScienceWorld, which provide dense task scores, we report both average score and success rate (\%).
All experiments are conducted three times, and the results are reported as the average value.

\subsubsection{Backbone LLMs.}
We report the main results with two backbone LLMs: Gemini-2.5-Pro~\citep{comanici2025gemini} and GPT-5.2-Codex~\citep{openai2025gpt52codex}. 
All additional experiments and analyses are conducted under Gemini-2.5-Pro.

\subsubsection{Baselines.}
For a fair comparison, we evaluate HiSkill against training-free methods from three categories. 
\textbf{Prompt-based methods} include \textbf{ReAct}, which interleaves reasoning, actions, and observations for step-by-step task solving~\citep{yao2023react}, and 
\textbf{Reflexion}, which converts task feedback into verbal reflections for self-improvement~\citep{shinn2023reflexion}. 
\textbf{Memory-based methods} include \textbf{ExpeL}, which extracts natural-language insights from past tasks and retrieves them during inference~\citep{zhao2024expel}; 
\textbf{Mem0}, which maintains a long-term experience pool to guide action selection~\citep{chhikara2025mem0}; 
\textbf{MemP}, which distills historical trajectories into procedural memory for decision making~\citep{fang2025memp}; and 
\textbf{SimpleMem}, which manages hierarchical memory for long-term information retrieval~\citep{liu2026simplemem}. 
\textbf{Skill-based methods} include \textbf{Vector Skills}, which performs flat semantic retrieval over skills; \textbf{SkillNet}, which builds a multi-task skill library for task-specific skill retrieval~\citep{liang2026skillnet}; and \textbf{GoS}, which retrieves dependency-aware skill bundles from an offline skill graph using semantic--lexical seeding and Personalized PageRank~\citep{liu2026graph}.

\subsubsection{Implementation Details.}

The implementation details of HiSkill are provided in Appendix~\ref{app:graph-construction-details}--\ref{app:subgraph-execution-details}. 
Besides, we tune two hyperparameters on the validation set: the dense--sparse retrieval weight $\lambda$ over $\{0,0.25,0.50,0.75,1.00\}$ and the seed node budget $K$ over $\{2,4,6,8,10\}$. 
Both analyses are reported in the Hyperparameter Analysis section.

\subsection{Main Results (RQ1)}
\label{sec:main-results}

Tables~\ref{tab:alfworld_seen_unseen} and~\ref{tab:webshop_scienceworld_results} report the results of our method and the baselines on three interactive environments with Gemini-2.5-Pro. The counterpart results with GPT-5.2-Codex are provided in Appendix~\ref{app:gpt52-results}. Overall, our method achieves the best overall performance on all datasets. 
Compared with the strongest baseline across all datasets and splits, our method achieves an average relative improvement of 17.33\% in success rate and 22.95\% in score. On ALFWorld, it obtains the best or tied-best results on 5 out of 6 subtasks, indicating stable advantages across different task types.

\subsubsection{Compared with Prompt-based Methods.}
ReAct and Reflexion rely mainly on in-context reasoning, actions, and feedback for step-by-step decision making. Although they perform well on some subtasks, they do not explicitly reuse historical interaction experience, leading to less stable performance. By contrast, our method abstracts reusable procedures from historical trajectories into a directed graph. On the ScienceWorld Seen split, compared with the strongest baseline Reflexion, our method improves score and success rate by 32.93\% and 31.45\%, respectively.

\subsubsection{Compared with Memory-based Methods.}
ExpeL, Mem0, MemP, and SimpleMem enhance decision making with historical experience, external memory, or procedural memory. However, their experience is typically stored as textual memories, summaries, or trajectory fragments, without explicitly organizing high-level procedures, executable actions, and execution relations. Compared with the strongest baselines, our method improves ALFWorld Seen and Unseen overall success rates over MemP by 7.32\% and 4.24\%, improves WebShop score and success rate over ExpeL by 11.14\% and 22.02\%, and improves ScienceWorld Unseen score and success rate over MemP by 25.30\% and 32.55\%.

\subsubsection{Compared with Skill-based Methods.}
Vector Skills, SkillNet, and GoS all exploit skill information, but they remain limited by flat semantic retrieval, skill-library selection, or skill-level dependency completion. These methods mainly use skills as high-level strategies or textual guidance, leaving a gap to executable actions. Our method jointly models skill nodes, AtomicOp nodes, and typed relations, and further uses task state for runtime decision making, which leads to consistently stronger performance than existing skill-based baselines across all three datasets.

\begin{table}[t]
\centering
\small
\setlength{\tabcolsep}{3.8pt}
\renewcommand{\arraystretch}{1.10}

\begin{tabular}{lccccc}
\toprule

\multirow{3}{*}{\textbf{Method}}
& \multicolumn{2}{c}{\textbf{ALFWorld}}
& \multirow{2}{*}{\textbf{WebShop}}
& \multicolumn{2}{c}{\textbf{ScienceWorld}} \\

\cmidrule(lr){2-3}
\cmidrule(lr){5-6}

& \textbf{Seen}
& \textbf{Unseen}
&
& \textbf{Seen}
& \textbf{Unseen} \\

\midrule

\rowcolor{gray!16}
\multicolumn{6}{l}{\textit{\textbf{Prompt-based Methods}}} \\

ReAct
& 27,324 & 28,315
& 133,164
& 32,140 & 30,055 \\

Reflexion
& 48,486 & 43,180
& 363,789
& \underline{22,752} & 25,678 \\

\midrule

\rowcolor{gray!16}
\multicolumn{6}{l}{\textit{\textbf{Memory-based Methods}}} \\

ExpeL
& \underline{15,173} & \underline{19,021}
& \underline{76,505}
& 24,246 & \underline{25,024} \\

Mem0
& 50,555 & 52,374
& 131,969
& 60,560 & 51,605 \\

MemP
& 32,476 & 34,110
& 156,291
& 53,224 & 60,297 \\

SimpleMem
& 26,053 & 27,880
& 135,448
& 60,571 & 81,527 \\

\midrule

\rowcolor{gray!16}
\multicolumn{6}{l}{\textit{\textbf{Skill-based Methods}}} \\

Vector Skills
& 52,973 & 62,280
& 139,953
& 42,101 & 43,485 \\

SkillNet
& 66,053 & 58,992
& 77,731
& 295,143 & 299,667 \\

GoS
& 58,118 & 59,407
& 138,240
& 41,006 & 46,867 \\

\textbf{HiSkill}
& \textbf{6,462} & \textbf{6,838}
& \textbf{4,501}
& \textbf{7,747} & \textbf{8,135} \\

\bottomrule

\end{tabular}

\caption{Average token consumption per task. Lower is better. The best results are highlighted in boldface, and the second-best results are underlined.}
\label{tab:token_cost}

\end{table}

\begin{table}[t]
\centering
\begingroup
\small
\setlength{\tabcolsep}{3.8pt}
\renewcommand{\arraystretch}{1.10}
\begin{tabular}{lccccc}
\toprule
\multirow{3}{*}{\textbf{Method}}
& \multicolumn{2}{c}{\textbf{ALFWorld}}
& \multirow{2}{*}{\textbf{WebShop}}
& \multicolumn{2}{c}{\textbf{ScienceWorld}} \\
\cmidrule(lr){2-3}
\cmidrule(lr){5-6}
& \textbf{Seen}
& \textbf{Unseen}
&
& \textbf{Seen}
& \textbf{Unseen} \\
\midrule

w/o Atom.
& 87.14 & 86.57
& 62.20
& 77.32 & 78.20 \\

w/o Edge
& 84.29 & 84.33
& 60.80
& 74.23 & 73.46 \\

w/o Sup./Rec.
& \underline{87.86} & \underline{89.55}
& \underline{63.80}
& 78.35 & 74.88 \\

w/o State
& 86.42 & 85.82
& 62.60
& \underline{79.38} & \underline{78.67} \\

Static Subgraph
& 75.00 & 79.10
& 53.00
& 69.59 & 68.72 \\

Top-$K$ Prompt
& 66.43 & 78.36
& 40.80
& 57.73 & 47.39 \\

\midrule

GoS
& 62.86 & 65.67
& 36.80
& 38.14 & 41.71 \\

OurG+GoSUse
& 80.00 & 82.86
& 53.20
& 51.03 & 54.50 \\

GoSG+OurUse
& 72.86 & 75.71
& 51.80
& 56.70 & 51.18 \\

\textbf{Full}
& \textbf{94.29} & \textbf{91.79}
& \textbf{67.60}
& \textbf{84.02} & \textbf{81.04} \\

\bottomrule
\end{tabular}
\endgroup
\caption{Ablation study results measured by success rate (\%). The best results are highlighted in boldface, and the second-best results are underlined.}
\label{tab:ablation_results}
\end{table}

\subsection{Token Consumption Analysis (RQ2)}
\label{sec:cost-analysis}

Table~\ref{tab:token_cost} compares the average token consumption per task. 
Our method consistently achieves the lowest token usage across all datasets and splits. 
Compared with the lowest-token baseline on each setting, HiSkill reduces token consumption by 61.10\%, 94.12\%, and 66.76\% on ALFWorld, WebShop, and ScienceWorld, respectively. 

This indicates that the performance gains of HiSkill do not come from providing the LLM with longer trajectories, larger memory contents, or extensive skill texts. 
Instead, the retrieved task-relevant subgraph provides a compact representation of reusable experience, enabling the LLM to reason over structured skills and executable actions with substantially reduced context, demonstrating that hierarchical skill graphs enable more efficient experience utilization during long-horizon task execution.

\subsection{Ablation Studies (RQ3)}
\label{sec:ablation}

Table~\ref{tab:ablation_results} reports the success rates of different ablation variants. 
Overall, the full method achieves the best results across all datasets and splits. 
Compared with the strongest ablated variant, it obtains relative improvements of 7.32\%, 2.50\%, 5.96\%, 5.85\%, and 3.01\% on ALFWorld Seen, ALFWorld Unseen, WebShop, ScienceWorld Seen, and ScienceWorld Unseen, respectively. 
The variants are analyzed as follows.

\subsubsection{Graph Representation.}
\textbf{w/o Atom.} removes AtomicOp nodes as independent nodes for retrieval and dispatch, while \textbf{w/o Edge} removes all typed edges. 
Both variants degrade performance, indicating that high-level skills alone are insufficient for fine-grained interactive execution. 
AtomicOps bridge the gap between skill descriptions and executable actions, while typed edges make the retrieved subgraph a usable procedural context rather than a set of isolated nodes.

\subsubsection{Runtime Mechanisms.}
\textbf{w/o Sup./Rec.} removes support and recovery edges and the corresponding inference-time operations, while \textbf{w/o State} removes the task state. 
Both variants underperform the full method, showing that support/recovery candidates are important for handling missing prerequisites and stalled execution, while the task state provides the runtime evidence required for reliable skill switching and AtomicOp selection.

\subsubsection{Subgraph Usage.}
\textbf{Static Subgraph} directly uses the retrieved subgraph as a fixed execution structure without using the LLM to dynamically select skills or AtomicOps, while 
\textbf{Top-$K$ Prompt} only injects the retrieved top-$K$ skill texts into the prompt. 
Their lower performance suggests that the retrieved subgraph is not a fixed execution plan, and that high-level skill text alone does not reliably  help the LLM infer concrete steps, action arguments, or execution timing.

\subsubsection{Component Transfer.}
\textbf{OurG+GoSUse} uses our graph representation with the GoS-style retrieval and skill-bundle usage strategy, while \textbf{GoSG+OurUse} uses the GoS graph with our subgraph retrieval and task execution mechanism. 
OurG+GoSUse clearly outperforms vanilla GoS, showing that our graph representation provides more useful hierarchical and typed structure. 
GoSG+OurUse also improves over GoS, indicating that our execution mechanism can better use an existing skill graph. 
However, both transfer variants remain below the full method, suggesting that graph representation and runtime usage should be designed jointly.

\begin{figure}[t]
  \centering
  \includegraphics[width=\linewidth]{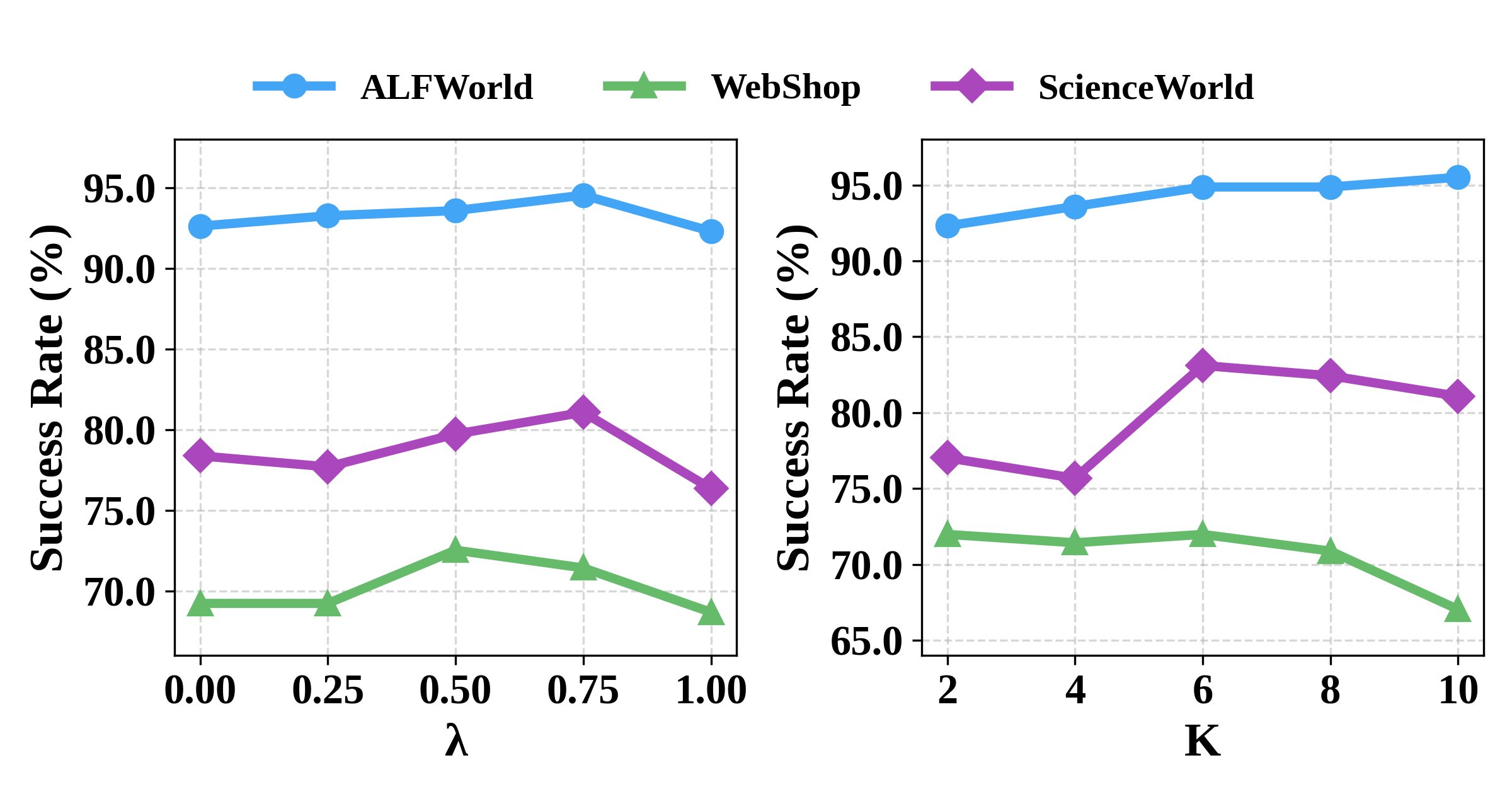}
  \caption{Validation sensitivity to the retrieval weight $\lambda$ and seed node budget $K$ measured by success rate.}
  \label{fig:hyper_success}
\end{figure}

\subsection{Hyperparameter Analysis (RQ4)}
\label{sec:hyperparameter_analysis}

We analyze two hyperparameters on the validation set: the dense--sparse retrieval weight $\lambda$ and the seed node budget $K$. 
Figure~\ref{fig:hyper_success} shows the results under different hyperparameters. 

For $\lambda$, intermediate values generally outperform using only dense or sparse retrieval, indicating that semantic similarity and lexical matching provide complementary signals. 
ALFWorld is relatively insensitive to $\lambda$, while WebShop performs best at $\lambda=0.5$ and ScienceWorld at $\lambda=0.75$. 
Therefore, we select the best $\lambda$ for each dataset based on validation performance and keep it fixed in the main experiments.

For the seed node budget $K$, increasing $K$ can provide more candidate skill and AtomicOp nodes, but also introduces more context and decision noise. 
The results show that $K=6$ provides a strong overall trade-off: it achieves the best or near-best success rate on the three datasets, while avoiding the larger context cost of higher budgets. 
Although larger $K$ slightly improves ALFWorld, it degrades WebShop and ScienceWorld. 
We therefore set $K=6$ in the main experiments. 
Score-based sensitivity results for WebShop and ScienceWorld are provided in Appendix~\ref{app:hyperparameter_score}.

\section{Conclusion}
\label{sec:conclusion}

In this paper, we introduce HiSkill, a hierarchical skill graph framework for augmenting LLM agents in interactive tasks. 
HiSkill organizes historical interaction trajectories into a directed graph with skill nodes, AtomicOp nodes, and typed edges.
At inference time, HiSkill retrieves a task-relevant subgraph and then monitors a task state and an active skill. Together with the subgraph and an LLM, it performs skill switching, AtomicOp selection, and action grounding.
Experimental results show that HiSkill consistently improves task performance while substantially reducing inference token consumption across interactive environments. 
Future work will explore online skill graph evolution, adaptive skill refinement, and broader applications to more open-ended interactive scenarios.

\bibliographystyle{conference}
\bibliography{conference}

\clearpage
\setcounter{secnumdepth}{2}
\appendix

\begin{center}
{\LARGE \textbf{Appendix}}
\end{center}

\section{Details of Hierarchical Skill Graph Construction}
\label{app:graph-construction-details}
This appendix provides implementation details for trajectory processing, AtomicOp extraction, skill discovery, and typed edge induction.

\subsection{Trajectory Processing}
\label{app:trajectory-processing-details}

The system interacts with environments and collects task-solving trajectories, where each trajectory contains the task description, observations, raw actions, environment feedback, and the final task outcome. 
We store each trajectory as:
\begin{equation} \begin{array}{l} \{ \mathrm{task\_id}, \mathrm{env\_name}, \mathrm{task\_description}, \mathrm{task\_type}, \mathrm{steps}, \mathrm{success}, \mathrm{metadata}\}. \end{array} \label{eq:operation-modes} \end{equation}
Here, $\mathtt{steps}$ contains the interaction history, including observations, raw actions, and feedback, while $\mathtt{success}$ indicates whether the task is completed. 
This unified format allows trajectories from different environments to be processed by the same construction pipeline.

\subsection{AtomicOp Extraction}
\label{app:atomicop-extraction-details}

Raw actions often have environment-specific surface forms and contain task-specific entities. 
Each raw action is parsed into an action type and structured arguments. 
Task-specific entities and constraints identified from the task description are replaced with semantic placeholders, while reusable literal values are retained.
The normalized template is serialized as
$\mathtt{action\_type\mid key=value\mid\cdots}$, with argument keys sorted to ensure a stable representation. 
Occurrences with the same normalized template are deduplicated and merged into one AtomicOp node. 
Each AtomicOp keeps the raw action for traceability, the normalized template for reuse, placeholder arguments for instantiation, and support statistics for later retrieval and graph construction. 
Figure~\ref{fig:atomicop-examples} shows representative canonicalization examples.

\subsection{Skill Discovery}
\label{app:skill-discovery-details}

We convert successful trajectories into AtomicOp sequences and enumerate recurring AtomicOp subsequences. 
A candidate subsequence is retained only if it appears in at least two successful trajectories and contains at least two AtomicOps. 
Duplicate or heavily overlapping candidates are further filtered to keep compact and distinct patterns.

For each retained pattern, we select a representative normalized AtomicOp sequence and abstract it into a skill node. 
When multiple candidate sequences are available, we prefer the one that is more general and easier to instantiate, particularly one that better preserves task-level placeholders such as target objects, target receptacles, tools, or product constraints. 
The selected sequence defines the executable procedure of the skill. 
From its supporting trajectories, we derive structured metadata, including pre-states, post-states, argument candidates, representative examples, support count, and failure hints. 
For each argument field, at most six candidate values are retained to keep the skill representation compact.

Figure~\ref{fig:skill-examples} presents representative skill records, showing their AtomicOp sequence, pre-states, post-states, examples, failure hints, and support count. 
Additional metadata, such as complete argument candidates, constraint tags, and detailed failure descriptions, is retained in the skill graph but omitted from the figure for readability.

\begin{figure}[t]
\centering

\begin{tcolorbox}[
    figurebox,
    colback=gray!3,
    colframe=gray!45,
    title=AtomicOp Canonicalization Examples
]

\begin{tcolorbox}[
    atomcard,
    colback=blue!4,
    colframe=blue!55,
    title=\texttt{go\_to}
]
\field{Raw Action}
\code{go to cart}

\field{Normalized Template}
\code{go\_to | location=\{target\_receptacle\}}

\field{Arguments}
\code{\{"location": "\{target\_receptacle\}"\}}

\field{Support}
\code{3436}
\end{tcolorbox}

\vspace{1mm}

\begin{tcolorbox}[
    atomcard,
    colback=green!4,
    colframe=green!55!black,
    title=\texttt{move}
]
\field{Raw Action}
\code{move dishsponge to cart}

\field{Normalized Template}
\code{move | object=\{target\_object\} | destination=\{target\_receptacle\}}

\field{Arguments}
\code{\{"object": "\{target\_object\}", "destination": "\{target\_receptacle\}"\}}

\field{Support}
\code{3394}
\end{tcolorbox}

\vspace{1mm}

\begin{tcolorbox}[
    atomcard,
    colback=orange!5,
    colframe=orange!70!black,
    title=\texttt{take}
]
\field{Raw Action}
\code{take toiletpaper from countertop}

\field{Normalized Template}
\code{take | object=\{target\_object\ | source=countertop}

\field{Arguments}
\code{\{"object": "\{target\_object\}", "source": "countertop"\}}

\field{Support}
\code{235}
\end{tcolorbox}

\end{tcolorbox}

\caption{Examples of AtomicOp canonicalization.}
\label{fig:atomicop-examples}

\end{figure}

\begin{figure}[t]
\centering

\begin{tcolorbox}[
    figurebox,
    colback=gray!3,
    colframe=gray!45,
    title=Skill Representation Examples
]
\begin{tcolorbox}[
skillcard,
colback=blue!4,
colframe=blue!55,
title=Pick Two Objects and Place
]
\skillfield{AtomicOp Sequence}
\skillstep{go\_to}$\rightarrow$
\skillstep{take}$\rightarrow$
\skillstep{go\_to}$\rightarrow$
\skillstep{move}$\rightarrow$
\skillstep{go\_to}$\rightarrow$
\skillstep{take}$\rightarrow$
\skillstep{go\_to}$\rightarrow$
\skillstep{move}
\par
\skillfield{Pre-states}
\skillcode{holding\_target, target\_located, navigation\_needed}
\par
\skillfield{Post-states}
\skillcode{location\_progress, target\_placed}
\par
\skillfield{Example}
\skillcode{\emph{put two toiletpaper in sidetable.}}
\par
\skillfield{Failure Hint}
\skillcode{Loop after only one target is found.}
\par
\skillfield{Support}
\skillcode{52}
\end{tcolorbox}

\vspace{0.5mm}

\begin{tcolorbox}[
skillcard,
colback=green!4,
colframe=green!55!black,
title=Clean and Place
]
\skillfield{AtomicOp Sequence}
\skillstep{go\_to}$\rightarrow$
\skillstep{take}$\rightarrow$
\skillstep{go\_to}$\rightarrow$
\skillstep{clean}$\rightarrow$
\skillstep{go\_to}$\rightarrow$
\skillstep{move}
\par
\skillfield{Pre-states}
\skillcode{target\_located, tool\_ready, navigation\_needed}
\par
\skillfield{Post-states}
\skillcode{process\_progress, target\_placed}
\par
\skillfield{Example}
\skillcode{\emph{clean some kettle and put it in cabinet.}}
\par
\skillfield{Failure Hint}
\skillcode{Repeat checked locations or miss tool cues.}
\par
\skillfield{Support}
\skillcode{330}

\end{tcolorbox}

\vspace{0.5mm}

\begin{tcolorbox}[
skillcard,
colback=orange!5,
colframe=orange!70!black,
title=Pick Cool then Place in Receptacle
]
\skillfield{AtomicOp Sequence}
\skillstep{go\_to}$\rightarrow$
\skillstep{take}$\rightarrow$
\skillstep{go\_to}$\rightarrow$
\skillstep{cool}$\rightarrow$
\skillstep{go\_to}$\rightarrow$
\skillstep{move}
\par
\skillfield{Pre-states}
\skillcode{target\_located, tool\_ready, navigation\_needed}
\par
\skillfield{Post-states}
\skillcode{process\_progress, target\_placed}
\par
\skillfield{Example}
\skillcode{\emph{cool some mug and put it in coffeemachine.}}
\par
\skillfield{Failure Hint}
\skillcode{Search loops without locating the target.}
\par
\skillfield{Support}
\skillcode{330}
\end{tcolorbox}
\end{tcolorbox}

\caption{Examples of skill representation.}
\label{fig:skill-examples}

\end{figure}

\paragraph{Skill Text Refinement.}
After a skill is discovered from recurring AtomicOp patterns, we use an LLM only to refine its natural-language fields for readability and retrieval. 
The refinement prompt rewrites the skill name, description, and failure hints, while keeping the AtomicOp sequence and structured metadata unchanged. 
This ensures that the skill structure remains grounded in trajectory evidence, and the LLM only improves the textual representation. 
Figure~\ref{fig:skill_refinement_prompt} shows the prompt template.

\begin{figure}[t]
\centering
\begin{tcolorbox}[
    promptbox,
    width=\linewidth
]
\begin{lstlisting}[
  style=promptstyle,
  numbers=none,
  linewidth=\linewidth,
  aboveskip=0.8em
]
You are given a skill discovered from recurring AtomicOp sequences.
Generate a concise textual representation grounded only in the provided information.

Requirements:
- Use a short and specific skill name.
- Write one concise paragraph describing the procedure and its applicability.
- Summarize one or two failure hints.
- Do not add, remove, or reorder AtomicOps.
- Do not invent arguments, states, or tools.
- Return valid JSON only.

Return JSON with keys:
name, description, failure_modes.

AtomicOp sequence:
{skill.action_signature}

Pre-states:
{skill.prestates}

Post-states:
{skill.poststates}

Argument candidates:
{skill.arg_candidates}

Examples:
{skill.examples}
\end{lstlisting}
\end{tcolorbox}
\vspace{3mm}
\caption{Prompt template for skill text refinement.}
\label{fig:skill_refinement_prompt}
\end{figure}

\subsection{Typed Edge Induction}
\label{app:typed-edge-induction-details}

We construct five types of edges, where each edge is represented as a directed triple $(u,v,\ell)$ with source node $u$, target node $v$, and edge type $\ell$. 
Here, skill nodes are denoted by $s\in V_S$, AtomicOp nodes by $o\in V_O$.
For statistical edges, we store a support count and a confidence score to indicate the strength of the construction evidence. 

\paragraph{Decomposition Edges.}
The $\mathtt{decomposes\_to}$ edge is directly derived from each discovered skill node to the AtomicOp nodes in its ordered procedure. 
For a skill node $s$ with procedure $(o_1,\ldots,o_n)$, we add $(s,o_k,\mathtt{decomposes\_to})$ and store the order index $k$. 
This edge specifies how a high-level skill is decomposed into executable AtomicOp templates.

\paragraph{Temporal Transition Edges.}
The $\mathtt{can\_follow}$ edge captures temporal continuation patterns in successful trajectories. 
We construct this relation at two granularities. 
For skill-to-skill edges, we count how often a skill node appears after another skill node in successful trajectories. 
For AtomicOp-to-AtomicOp edges, we count adjacent AtomicOp transitions. 
The confidence is the relative outgoing frequency:
\begin{equation}
\mathrm{conf}(u,v)
=
\frac{\mathrm{count}(u\rightarrow v)}
{\sum_{v'} \mathrm{count}(u\rightarrow v')}.
\end{equation}
For each source skill node, we keep at most $3$ following skill nodes. 
For each source AtomicOp node, we keep at most $2$ following AtomicOp nodes, requiring $\mathrm{count}\geq 2$ and $\mathrm{conf}\geq 0.2$. 
This edge provides temporally plausible next skill or AtomicOp candidates.

\paragraph{Compatibility Edges.}
The $\mathtt{compatible\_with}$ edge is constructed only between skill nodes. 
It is induced from skill co-occurrence in successful trajectories and does not require the two skill nodes to be adjacent. 
If two skill nodes frequently appear in the same successful task context and their pre/post-states do not conflict, we add symmetric directed edges between them. 
The confidence is computed as the relative co-occurrence frequency:
\begin{equation}
\mathrm{conf}(s_i,s_j)
=
\frac{\mathrm{count}(s_i,s_j)}
{\sum_{s'} \mathrm{count}(s_i,s')}.
\end{equation}
We retain at most $4$ compatible targets for each source skill node and require the co-occurrence count to be at least $2$. 
This edge indicates which skills can jointly serve the same task or subgoal.

\paragraph{Support Edges.}
We infer this relation from the local action context of skill nodes in successful trajectories. 
For each trajectory segment matched to a skill node $s$, we examine the immediately preceding AtomicOp node $o$. 
An AtomicOp node is considered a support candidate only when its post-states overlap with the pre-states of the skill node:
\begin{equation}
\mathrm{conf}(o,s)
=
J(\mathrm{post}(o),\mathrm{pre}(s)),
\end{equation}
where $J(\cdot,\cdot)$ denotes Jaccard overlap over state predicates. 
We accumulate counts over all matched trajectory segments, keep support edges with count at least $2$ and confidence at least $0.45$, and retain at most $4$ support AtomicOp nodes for each skill node. 
This edge provides auxiliary operations when the current state does not satisfy the prerequisites of a skill node, such as opening a container, selecting an option, or obtaining a required tool.

\paragraph{Recovery Edges.}
The $\mathtt{recovers\_with}$ edge connects a recovery skill/AtomicOp node to a failed or stalled skill/AtomicOp node. 
We infer this relation from two sources.

First, we extract direct recovery evidence from failed trajectories. 
When a trajectory contains a repeated ineffective AtomicOp pattern followed by a different AtomicOp pattern, e.g., $o_a,o_a,o_b$, we treat $o_b$ as a recovery candidate for $o_a$. 
This yields an AtomicOp-level edge $o_b \rightarrow o_a$. 
If the two AtomicOp nodes belong to different skill nodes, we also add a skill-level recovery edge between the corresponding skill nodes.

Second, we infer semantic recovery edges from failure evidence associated with failed or stalled source nodes. 
Here, failure evidence refers to the failure hints, repeated failed action, missing state predicate, or non-progress signal extracted from failed trajectories. 
For each source node $u$, we build a recovery query $x^{\mathrm{rec}}_u$ from this evidence and retrieve candidate skill/AtomicOp nodes. 
For each candidate node $v$, we compute three matching signals. 
\begin{equation}
\begin{gathered}
\mathrm{score}_{\mathrm{rec}}(v\mid u) = 2\,\mathrm{Dense}(x^{\mathrm{rec}}_u,d_v)
+\mathrm{Sparse}(x^{\mathrm{rec}}_u,d_v) + \mathrm{Sim}_{\mathrm{state}}(u,v),
\end{gathered}
\label{eq:recovery-score}
\end{equation}
where $\mathrm{Dense}(\cdot)$ and $\mathrm{Sparse}(\cdot)$ follow the definitions in Eq.~\eqref{eq:node-retrieval-score}. 
We assign a larger weight to $\mathrm{Dense}(\cdot)$ because semantic similarity is the primary signal for finding recovery nodes that are relevant to the failure evidence.
The state compatibility term is computed as:
\begin{equation}
\mathrm{Sim}_{\mathrm{state}}(u,v)
=
J(\mathrm{pre}(u),\mathrm{pre}(v))
+
J(\mathrm{post}(u),\mathrm{post}(v)).
\label{eq:recovery-state}
\end{equation}
A semantic recovery edge is retained only when the final recovery score is at least $0.20$. 
For each target node, we keep at most $4$ incoming recovery edges.
This edge is used when execution fails, stagnates, or enters repeated ineffective actions.

The key distinction between $\mathtt{supports}$ and $\mathtt{recovers\_with}$ lies in their triggering condition. 
$\mathtt{supports}$ is used before executing a skill when missing prerequisites are detected, while $\mathtt{recovers\_with}$ is used after failure, stagnation, or repeated ineffective behavior has been observed.

\begin{figure}[t]
\centering
\begin{tcolorbox}[
    promptbox,
    width=\linewidth
]
\begin{lstlisting}[
  style=promptstyle,
  numbers=none,
  linewidth=\linewidth,
  aboveskip=0.8em
]
You are selecting the active skill node for HiSkill execution.
Choose exactly one skill node from the provided candidates.
Do not create new skill nodes.

Rules:
- For initial selection, rerank the provided top-2 skill nodes.
- If the current skill is finished, select only from candidates connected by can_follow or compatible_with.
- If execution is stuck or repeatedly fails, select only from candidates connected by recovers_with.
- Return valid JSON only.

Return JSON with keys:
selected_skill_id, confidence, reason.

Task:
{task_description}

Current observation:
{current_observation}

Task state:
{task_state}

Active skill:
{active_skill}

Candidate skill nodes:
{candidate_skill_nodes}
\end{lstlisting}
\end{tcolorbox}
\vspace{3mm}
\caption{Prompt template for initial skill selection and runtime skill switching.}
\label{fig:skill-selection-prompt}
\end{figure}

\begin{figure}[t]
\centering
\begin{tcolorbox}[
    promptbox,
    width=\linewidth
]
\begin{lstlisting}[
  style=promptstyle,
  numbers=none,
  linewidth=\linewidth,
  aboveskip=0.8em
]
You are selecting the next AtomicOp for HiSkill execution.
Choose one AtomicOp node or null from the provided candidates.
Do not create new AtomicOp nodes.

Rules:
- In the default case, select the first pending AtomicOp of the active skill.
- If prerequisites are missing, select only from candidates connected by supports.
- If execution is stuck or repeatedly fails, select only from candidates connected by recovers_with.
- If no suitable AtomicOp exists, generate an action directly.
- Return valid JSON only.

Return JSON with keys:
selected_op_id or null, confidence, reason.

Task:
{task_description}

Current observation:
{current_observation}

Task state:
{task_state}

Active skill:
{active_skill}

Executed AtomicOps:
{executed_atomicops}

Candidate AtomicOp nodes:
{candidate_atomicop_nodes}
\end{lstlisting}
\end{tcolorbox}
\vspace{3mm}
\caption{Prompt template for AtomicOp selection.}
\label{fig:atomicop-selection-prompt}
\end{figure}

\begin{figure}[t]
\centering
\begin{tcolorbox}[
    promptbox,
    width=\linewidth
]
\begin{lstlisting}[
  style=promptstyle,
  numbers=none,
  linewidth=\linewidth,
  aboveskip=0.8em
]
You are the final HiSkill action grounder; please produce one concrete executable action.

Rules:
- If the target source is AtomicOp, instantiate the normalized action template using the provided argument candidates and evidence from the task, current observation, and task state.
- Keep the normalized template structure unchanged.
- If the target source is LLM direct-action, AtomicOp instantiation is bypassed.
- Follow the environment action syntax.
- Return valid JSON only.

Return JSON with keys:
action, arguments.

Task:
{task_description}

Current observation:
{current_observation}

Task state:
{task_state}

Selected AtomicOp:
{selected_atomicop_or_none}

Normalized action template:
{normalized_action_template}

Argument candidates:
{argument_candidates}
\end{lstlisting}
\end{tcolorbox}
\vspace{3mm}
\caption{Prompt template for action grounding.}
\label{fig:action-grounding-prompt}
\end{figure}

\section{Details of Task-relevant Skill Subgraph Retrieval}
\label{app:subgraph-retrieval-details}

This section provides implementation details of task-relevant skill subgraph retrieval, including seed node retrieval and graph hydration.

\subsection{Seed Node Retrieval}
\label{app:seed-node-retrieval-details}

For seed node retrieval, we first convert both the task and graph nodes into textual representations before computing relevance scores. 
Specifically, the retrieval query $x_q$ is constructed by concatenating the task description and the initial observation. 
For each graph node $v$, we construct a retrieval text $d_v$ by concatenating multiple node attributes while skipping empty fields, removing duplicated text segments, and compressing redundant spaces.
For skill nodes, $d_v$ contains the skill name, description, pre-states, post-states, examples and failure hints. 
For AtomicOp nodes, $d_v$ contains the normalized action template and argument candidates. 

The dense similarity in Eq.~\eqref{eq:node-retrieval-score} is computed using text-embedding-3-large~\citep{openai2024embedding}, where the query and node texts are encoded into embeddings and ranked by cosine similarity. 
The sparse score is computed using BM25~\citep{robertson2009probabilistic} based on lexical overlap between the query and node text.

Before combining the two retrieval signals, we independently normalize the dense and sparse scores using min-max normalization:
\begin{equation}
\mathrm{norm}(s_v)
=
\frac{s_v-\min(s)}
{\max(s)-\min(s)},
\end{equation}
where $s_v$ denotes the raw dense or sparse score of node $v$, and $s$ denotes the set of raw scores from all candidate nodes under the corresponding retrieval signal.
When all candidate nodes have identical scores, we set all normalized scores to $1.0$ if the original score is positive and to $0.0$ otherwise.
This normalization is applied separately to dense and sparse scores to avoid scale differences between embedding similarity and BM25 scores.

For BM25 indexing, all node texts are tokenized using a unified preprocessing procedure. 
Specifically, text is converted to lowercase and only alphabetic and numeric tokens are retained using regular expression filtering. 
The index statistics, including term frequency, text frequency, text length, and average text length, are computed over all node texts and used for BM25 scoring.

Finally, skill nodes and AtomicOp nodes are ranked separately according to Eq.~\eqref{eq:node-retrieval-score}. 
The top-$K$ nodes are selected as retrieval seed nodes, and their union forms the seed node set $V_{\mathrm{seed}}$.

\subsection{Graph Hydration}
\label{app:graph-hydration-details}

After seed node retrieval, we expand the retrieved nodes using the typed relations to form the task-relevant subgraph. 
Specifically, for each selected skill node, we include the AtomicOp nodes connected by $\mathtt{decomposes\_to}$ edges to preserve its executable action templates.
For the remaining typed edges, we perform one-hop neighbor expansion by including nodes connected to the selected nodes.
Candidate neighbors are ranked using Eq.~\eqref{eq:node-retrieval-score}. 
To control the size of the retrieved subgraph, we keep the top-2 neighbors for each non-decomposition relation.
Finally, we retain all typed edges among the selected nodes to form the task-relevant subgraph $G_q=(V_q,E_q)$.

\section{Details of Subgraph-guided Task Execution}
\label{app:subgraph-execution-details}

This appendix provides implementation details of skill selection and switching, AtomicOp selection, and action grounding.

\subsection{Skill Selection and Switching}
\label{app:skill-switching-details}

\paragraph{Initial Skill Selection.}
The skill nodes in $V_q$ are ranked according to the retrieval score in Eq.~\eqref{eq:node-retrieval-score}. 
The top-2 skill nodes are provided to the LLM, which reranks them according to the task description, initial observation, and task state, and selects the more suitable node as the initial active skill.

\paragraph{Runtime Skill Switching.}
The active skill is switched according to the current task state and the typed relations in $G_q$. 
When the active skill is finished, successor candidates are collected through $\mathtt{can\_follow}$ or $\mathtt{compatible\_with}$ edges. When execution is stuck or repeatedly fails, recovery candidates are collected through $\mathtt{recovers\_with}$ edges.

The prompt templates for initial skill selection and runtime skill switching are shown in Figure~\ref{fig:skill-selection-prompt}.

\subsection{AtomicOp Selection}
\label{app:atomicop-selection-details}

AtomicOp selection is conditioned on the active skill, current task state, and typed relations in $G_q$. 
When a skill becomes active, the controller tracks its AtomicOp execution progress according to the order indices of the $\mathtt{decomposes\_to}$ edges, marking each sequence position as executed or pending.
By default, the first pending AtomicOp of the active skill is selected. 
When prerequisites are missing, support candidates are collected through $\mathtt{supports}$ edges. 
When execution is stuck or repeatedly fails, recovery candidates are collected through $\mathtt{recovers\_with}$ edges. 
If no candidate is suitable, the system enables an LLM direct-action.

The prompt template for AtomicOp selection is shown in Figure~\ref{fig:atomicop-selection-prompt}.

\subsection{Action Grounding}
\label{app:action-grounding-details}

When the selected target is an AtomicOp node, the system instantiates its normalized action template. 
Placeholder values are resolved using the AtomicOp argument candidates together with evidence from the task description, current observation, and task state. 
The instantiated template is then converted into a concrete action that follows the environment action syntax.

When the selected target is an LLM direct-action, AtomicOp instantiation is bypassed, and the LLM directly generates one concrete action.
Figure~\ref{fig:action-grounding-prompt} shows the prompt template.

\begin{table}[t]
\centering
\scriptsize
\setlength{\tabcolsep}{1.8pt}
\renewcommand{\arraystretch}{1.05}

\resizebox{\linewidth}{!}{
\begin{tabular}{lccccc cccc ccc}
\toprule
\multirow{2}{*}{\textbf{Dataset}}
& \multicolumn{2}{c}{\textbf{Traj.}}
& \multicolumn{3}{c}{\textbf{Split}}
& \multicolumn{4}{c}{\textbf{Skill Graph}}
& \multicolumn{3}{c}{\textbf{Retrieved Subgraph}} \\
\cmidrule(lr){2-3}
\cmidrule(lr){4-6}
\cmidrule(lr){7-10}
\cmidrule(lr){11-13}

& \textbf{Succ.}
& \textbf{Fail.}
& \textbf{Val.}
& \textbf{Seen}
& \textbf{Unseen}
& \textbf{Ops}
& \textbf{Skills}
& \textbf{Edges}
& \textbf{Ops/Skl.}
& \textbf{Ops}
& \textbf{Skills}
& \textbf{Edges} \\

\midrule

\textbf{ALFWorld}     & 1262 & 1545 & 312 & 140 & 134 & 112 & 66 & 649 & 3.8 & 11.1 & 8.1 & 70.2 \\
\textbf{WebShop}      & 646  & 996  & 182 & 500 & --  & 134 & 69 & 608 & 2.8 & 9.2  & 7.7 & 66.2 \\
\textbf{ScienceWorld} & 800  & 535  & 148 & 194 & 211 & 134 & 67 & 573 & 2.7 & 13.7 & 9.3 & 62.4 \\

\bottomrule
\end{tabular}
}

\caption{Statistics of datasets.}
\label{tab:dataset-graph-stats}

\end{table}

\section{Dataset Details}
\label{app:dataset-details}

This section provides additional details about the three interactive datasets used in our experiments. 
Table~\ref{tab:dataset-graph-stats} summarizes the trajectory statistics, evaluation splits, constructed hierarchical skill graph sizes, and average retrieved subgraph sizes.

\paragraph{ALFWorld.}
ALFWorld~\citep{shridhar2020alfworld} is a text-based embodied environment aligned with the ALFRED benchmark. 
In each episode, the system receives a textual goal and completes it through multi-turn interaction with the environment using text commands. 
Following prior work~\citep{feng2025group}, ALFWorld contains six common household task categories: Pick and Place (Pick), Examine in Light (Look), Clean and Place (Clean), Heat and Place (Heat), Cool and Place (Cool), and Pick Two and Place (Pick2).

\paragraph{WebShop.}
WebShop~\citep{yao2022webshop} is a web-based interactive shopping environment designed to evaluate sequential decision making in realistic online shopping scenarios. 
Given a user instruction, the system interacts with a shopping website through search, browsing, comparison, option selection, and purchase actions. 
The environment requires understanding compositional textual constraints and maintaining progress across multiple interaction steps.

\paragraph{ScienceWorld.}
ScienceWorld~\citep{wang2022scienceworld} is a text-based scientific experimentation environment for evaluating grounded scientific reasoning and interactive task completion. 
The system must observe the environment, locate relevant objects, operate tools or instruments, and use environment feedback to complete science-oriented tasks. 
The environment emphasizes long-horizon decision making, state tracking, and procedural control under partial observations.

\paragraph{Statistics.}
Table~\ref{tab:dataset-graph-stats} reports the statistics used in our experiments. 
The Success and Failure columns denote trajectories used for  hierarchical skill graph construction after holding out the validation split. 
The Val. column denotes the 10\% validation split held out from training data. 
The Skill Graph columns report the number of AtomicOp nodes, skill nodes, typed edges, and the average number of AtomicOps per skill. 
The Retrieved Subgraph columns report the average number of AtomicOps, skills, and edges in task-relevant retrieved skill subgraphs.

\begin{table}[t]
\centering
\scriptsize
\setlength{\tabcolsep}{1.0pt}
\renewcommand{\arraystretch}{1.08}

\resizebox{\linewidth}{!}{
\begin{tabular}{lccccccc|ccccccc}
\toprule
\multirow{3}{*}{\textbf{Method}}
& \multicolumn{14}{c}{\textbf{ALFWorld}}\\
\cmidrule(lr){2-15}

& \multicolumn{7}{c}{\textbf{Seen}}
& \multicolumn{7}{c}{\textbf{Unseen}}\\

\cmidrule(lr){2-8}
\cmidrule(lr){9-15}

& \textbf{Pick} & \textbf{Look} & \textbf{Clean} & \textbf{Heat} & \textbf{Cool} & \textbf{Pick2} & \textbf{All}
& \textbf{Pick} & \textbf{Look} & \textbf{Clean} & \textbf{Heat} & \textbf{Cool} & \textbf{Pick2} & \textbf{All} \\
\midrule

\rowcolor{gray!16}
\multicolumn{15}{l}{\textit{\textbf{Prompt-based Methods}}} \\

ReAct
& 94.29 & 92.31 & 74.07 & 43.75 & 56.00 & 79.17 & 75.00
& 95.83 & 94.44 & 87.10 & 47.83 & 76.19 & 82.35 & 80.60 \\
Reflexion
& \underline{97.14} & 69.23 & \underline{96.30} & \underline{87.50} & 76.00 & 75.00 & 85.71
& \textbf{100.00} & 83.33 & 80.65 & \underline{73.91} & 90.48 & 58.82 & 82.09 \\

\midrule
\rowcolor{gray!16}
\multicolumn{15}{l}{\textit{\textbf{Memory-based Methods}}} \\
ExpeL
& 91.43 & 92.31 & \underline{96.30} & 68.75 & 60.00 & 79.17 & 82.14
& 79.17 & 77.78 & 87.10 & 65.22 & 57.14 & 64.71 & 73.13 \\
Mem0
& \underline{97.14} & 92.31 & 92.59 & \underline{87.50} & 76.00 & \underline{95.83} & 90.71
& \textbf{100.00} & 66.67 & 90.32 & 56.52 & 95.24 & 82.35 & 82.84 \\
MemP
& \underline{97.14} & \textbf{100.00} & \textbf{100.00} & \underline{87.50} & \underline{92.00} & 91.67 & \textbf{95.00}
& \textbf{100.00} & \textbf{100.00} & \textbf{93.55} & \textbf{82.61} & \textbf{100.00} & 76.47 & \textbf{92.54} \\
SimpleMem
& \underline{97.14} & \textbf{100.00} & 62.96 & 43.75 & 68.00 & 87.50 & 77.86
& 95.83 & 83.33 & 80.65 & 26.09 & 52.38 & 70.59 & 68.66 \\

\midrule

\rowcolor{gray!16}
\multicolumn{15}{l}{\textit{\textbf{Skill-based Methods}}} \\
Vector Skills
& \underline{97.14} & 84.62 & 77.78 & 56.25 & 56.00 & 83.33 & 77.86
& \textbf{100.00} & 88.89 & 77.42 & 47.83 & 61.90 & 82.35 & 76.12 \\
SkillNet
& \underline{97.14} & 76.92 & 88.89 & \textbf{93.75} & 88.00 & \textbf{100.00} & 92.14
& 91.67 & 72.22 & \textbf{93.55} & 65.22 & 90.48 & \textbf{100.00} & 85.82 \\
GoS
& 94.29 & 92.31 & 81.48 & 31.25 & 40.00 & 62.50 & 69.29
& 87.50 & 88.89 & 77.42 & 34.78 & 52.38 & 70.59 & 68.66 \\
\textbf{HiSkill}
& \textbf{100.00} & \textbf{100.00} & 92.59 & 75.00 & \textbf{100.00} & \underline{95.83} & \textbf{95.00}
& \textbf{100.00} & \textbf{100.00} & 90.32 & 69.57 & \textbf{100.00} & \textbf{100.00} & \textbf{92.54} \\

\bottomrule
\end{tabular}
}

\caption{Performance on ALFWorld with GPT-5.2-Codex: Success rate (\%) is reported. The best results are highlighted in boldface, and the second-best results are underlined.}
\label{tab:alfworld_gpt}

\end{table}

\section{Additional Experimental Results}
\subsection{Main Results on GPT-5.2-Codex}
\label{app:gpt52-results}

Tables~\ref{tab:alfworld_gpt} and~\ref{tab:webshop_gpt} report the results with GPT-5.2-Codex. The overall trend is consistent with the Gemini-2.5-Pro results: our method remains the strongest or tied-strongest approach across the three environments, showing that the gains of HiSkill are consistent across different backbone LLMs. 
Compared with the strongest baseline across all datasets and splits, our method achieves an average relative improvement of 11.91\% in success rate and 18.31\% in score.
On ALFWorld, our method matches the best baseline on both Seen and Unseen overall success rates, where performance is already close to saturated. On WebShop, compared with the strongest baseline, our method achieves relative improvements of 26.64\% in score and 34.98\% in success rate. On ScienceWorld, the relative improvements over the strongest baseline are 18.91\% / 19.00\% on Seen and 11.15\% / 21.28\% on Unseen in score / success rate. These results confirm that explicitly organizing skill nodes, AtomicOp nodes, and typed relations is beneficial beyond a single backbone, especially in environments requiring fine-grained action grounding and long-horizon progress control.

\begin{table}[t]
\centering
\small
\setlength{\tabcolsep}{1.7mm}
\renewcommand{\arraystretch}{1.10}
\begin{tabular}{lcc|cc|cc}
\toprule
\multirow{3}{*}{\textbf{Method}}
& \multicolumn{2}{c|}{\textbf{WebShop}}
& \multicolumn{4}{c}{\textbf{ScienceWorld}} \\
\cmidrule(lr){2-3} \cmidrule(lr){4-7}
& \multirow{2}{*}{\textbf{Score}}
& \multirow{2}{*}{\textbf{Succ.}}
& \multicolumn{2}{c|}{\textbf{Seen}}
& \multicolumn{2}{c}{\textbf{Unseen}} \\
\cmidrule(lr){4-5} \cmidrule(lr){6-7}
&
&
& \textbf{Score}
& \textbf{Succ.}
& \textbf{Score}
& \textbf{Succ.} \\
\midrule

\rowcolor{gray!16}
\multicolumn{7}{l}{\textit{\textbf{Prompt-based Methods}}} \\
ReAct
& 54.11 & 42.40
& 68.03 & 63.92
& 58.79 & 54.98 \\
Reflexion
& 51.75 & \underline{48.60}
& 66.55 & 59.28
& 54.18 & 45.97 \\

\midrule
\rowcolor{gray!16}
\multicolumn{7}{l}{\textit{\textbf{Memory-based Methods}}} \\
ExpeL
& 35.71 & 31.40
& 60.34 & 54.12
& 64.02 & 53.55 \\
Mem0
& 57.83 & 46.20
& 73.50 & 69.07
& 67.46 & 64.93 \\
MemP
& \underline{59.24} & 48.00
& \underline{77.21} & \underline{73.20}
& \underline{75.43} & \underline{66.82} \\
SimpleMem
& 48.62 & 38.00
& 64.41 & 67.53
& 55.55 & 58.29 \\

\midrule
\rowcolor{gray!16}
\multicolumn{7}{l}{\textit{\textbf{Skill-based Methods}}} \\
Vector Skills
& 58.11 & 46.80
& 64.45 & 63.40
& 55.07 & 55.45 \\
SkillNet
& 27.76 & 23.60
& 65.32 & 64.43
& 56.64 & 60.19 \\
GoS
& 56.21 & 48.20
& 70.26 & 67.01
& 63.06 & 61.14 \\
\textbf{HiSkill}
& \textbf{75.02} & \textbf{65.60}
& \textbf{91.81} & \textbf{87.11}
& \textbf{83.84} & \textbf{81.04} \\

\bottomrule
\end{tabular}
\caption{Performance on WebShop and ScienceWorld with GPT-5.2-Codex: Average score and success rate (\%) are reported. The best results are highlighted in boldface, and the second-best results are underlined.}
\label{tab:webshop_gpt}
\end{table}

\subsection{Score-based Hyperparameter Analysis}
\label{app:hyperparameter_score}

\begin{figure}[t]
  \centering
  \includegraphics[width=\linewidth]{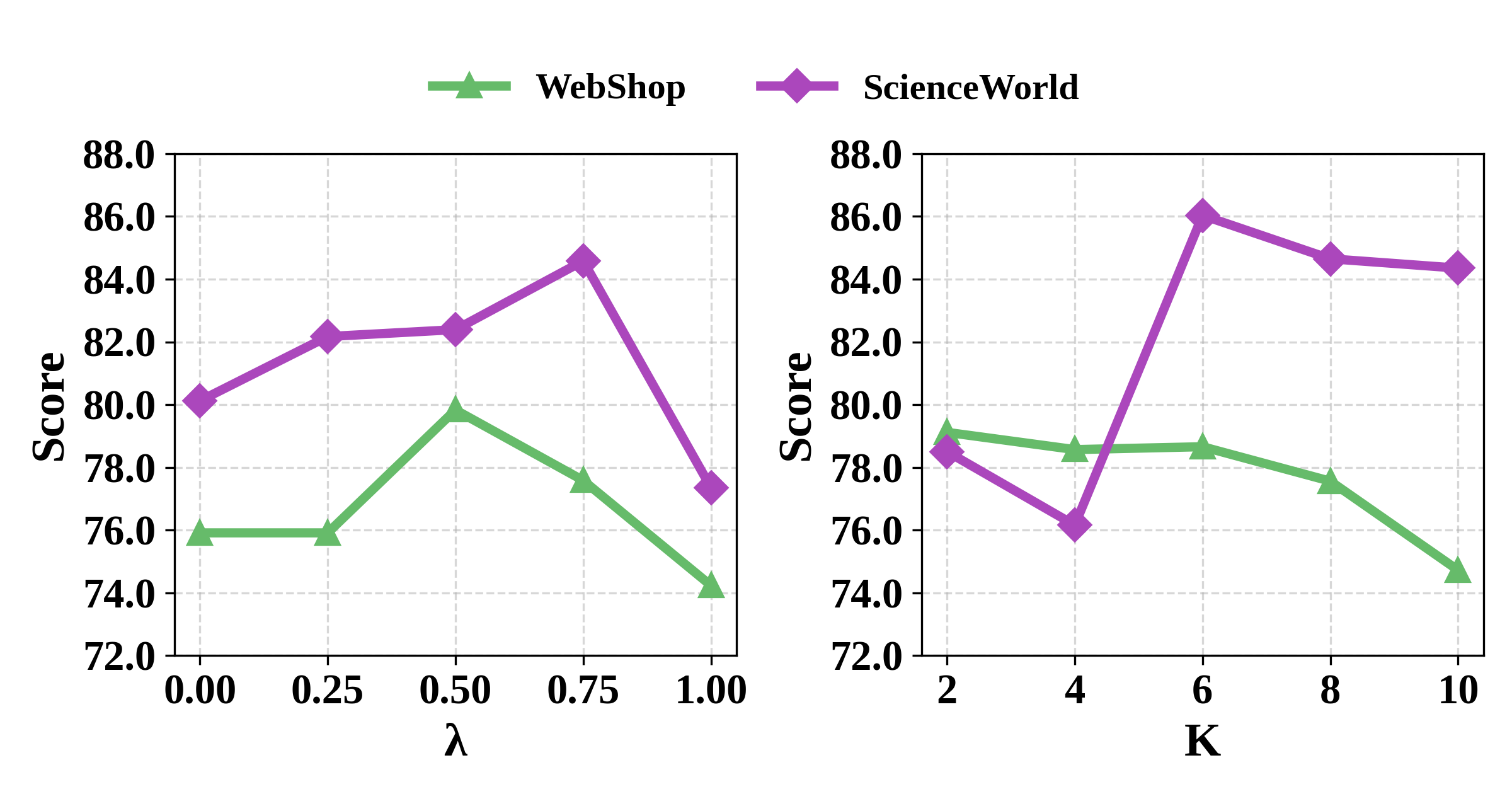}
  \caption{Validation sensitivity to the retrieval weight $\lambda$ and seed node budget $K$ measured by task score.}
  \label{fig:hyper_score}
\end{figure}

Figure~\ref{fig:hyper_score} reports the validation score under different retrieval weights $\lambda$ and seed node budgets $K$ on WebShop and ScienceWorld. 
ALFWorld is omitted because it only provides a binary success signal. 
The score trends are consistent with the success rate analysis in Figure~\ref{fig:hyper_success}: WebShop favors a balanced dense--sparse setting around $\lambda=0.5$, while ScienceWorld benefits more from semantic retrieval around $\lambda=0.75$. 
For the seed node budget, $K=6$ achieves the best ScienceWorld score and remains competitive on WebShop, supporting its use as the default budget in the main experiments.

\clearpage

\subsection{Case Study}
\label{app:case-study}

We present a WebShop case to illustrate how HiSkill retrieves a task-relevant subgraph, monitors a task state and an active skill, and leverages the retrieved subgraph with an LLM for skill switching, AtomicOp selection, and action grounding.

\begin{tcolorbox}[casepanel,title={Task and Initial Context}]

\begin{tabularx}{\linewidth}{
  @{}>{\raggedright\arraybackslash\bfseries}p{0.18\linewidth}
  >{\raggedright\arraybackslash}X@{}
}
Task $q$ &
Find me \textcolor{decisionred}{wash cold, machine wash men's shirts}, classic fit with
\textcolor{decisionred}{color: black}, and
\textcolor{decisionred}{fit type: youth}, and
\textcolor{decisionred}{size: 4t}, and \textcolor{decisionred}{price lower than 50.00 dollars}.

\\[0.6em]

Observation $o_0$ &
Search page.
\end{tabularx}

\vspace{0.4em}
\textbf{Task State $z_0$:}

\begin{lstlisting}[style=casestate]
stage: search
current_product: none
selected_options: []
search_history: []
visited_products: []
rejected_products: []
visited_navigation: []
recent_actions: []
recent_failures: []
@@missing_constraints: [product, color=black, fit_type=youth, size=4t]@@
\end{lstlisting}

\end{tcolorbox}

\begin{tcolorbox}[casepanel,title={Skill Subgraph Retrieval}]

\begin{minipage}[t]{0.49\linewidth}
\textbf{Seed Skill Nodes}
\begin{enumerate}
  \small
  \setlength{\itemsep}{0.15em}
  \item \texttt{s3}: SearchApparelWithConstraints
  \item \texttt{s4}: FilterAndBuyApparelByAttributes
  \item \texttt{s6}: SelectAndBuyApparelVariant
  \item \texttt{s9}: ApparelVariantSelectAndBuyNow
  \item \texttt{s7}: SelectSizeColorAndBuyNow
  \item \texttt{s8}: ApparelColorSizeBuy
\end{enumerate}
\end{minipage}
\hfill
\begin{minipage}[t]{0.49\linewidth}
\textbf{Seed AtomicOp Nodes}
\begin{enumerate}
  \small
  \setlength{\itemsep}{0.15em}
  \item \texttt{o10}: inspect
  \item \texttt{o7}: pick
  \item \texttt{o6}: click\_product()
  \item \texttt{o2}: select\_color()
  \item \texttt{o3}: select\_size()
  \item \texttt{o4}: buy()
\end{enumerate}
\end{minipage}

\vspace{0.4em}
Expanded nodes: \texttt{s1}: ApparelSearchSelectBuy, \texttt{s5}: SelectProductASIN, \texttt{o0}: search(), \texttt{o1}: click(), \texttt{o8}: submit.

After graph hydration, the task-relevant subgraph contains
$8$ skill nodes, $9$ AtomicOp nodes, and $66$ typed edges.

\end{tcolorbox}

\begin{tcolorbox}[
  casepanel,
  title={Task-relevant Subgraph $G_q$}
]
\centering
\includegraphics[
  width=0.85\linewidth,
  keepaspectratio
]{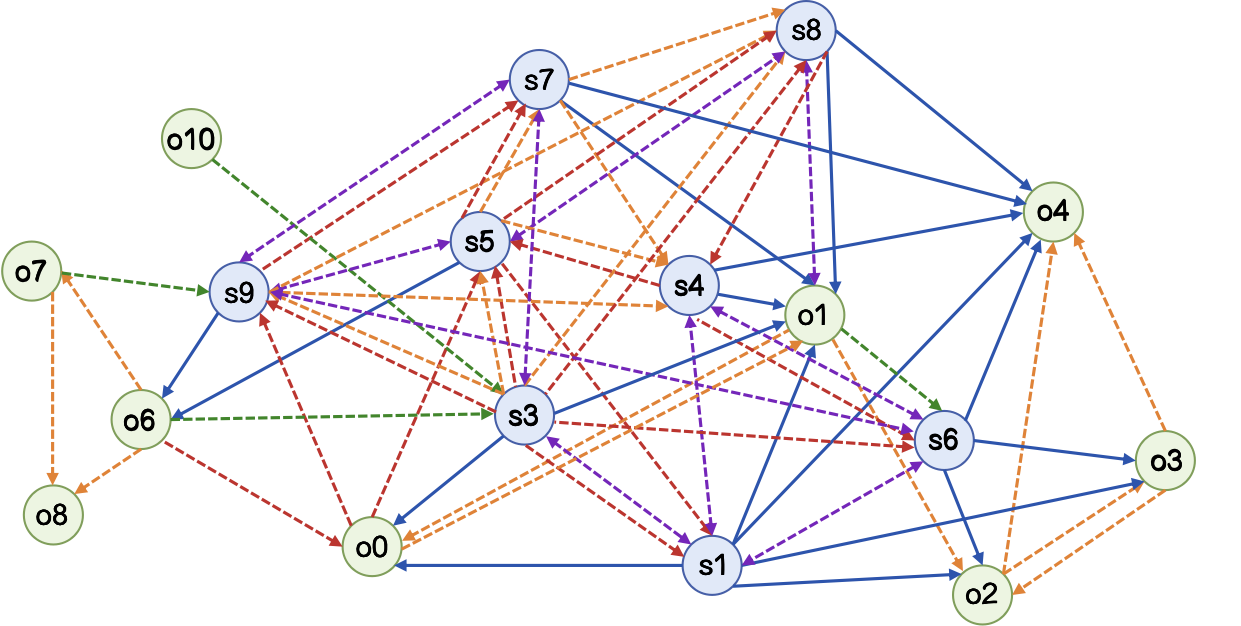}
\end{tcolorbox}

\begin{tcolorbox}[casestep,title={Step 0}]

\begin{tabularx}{\linewidth}{
  @{}>{\raggedright\arraybackslash\bfseries}p{0.20\linewidth}
  >{\raggedright\arraybackslash}X@{}
}

\casefield{Active Skill Node}{
  \texttt{s1}
}

\casefield{Subgraph Decision}{
  Select the initial skill node and follow
  $
  \texttt{s1}
  \xrightarrow{\mathtt{decomposes\_to}}
  \texttt{o0}
  $.
}

\casefield{Selected Target}{
  \texttt{o0}: search(query=men cold machine wash shirts black)
}

\casefield{Grounded Action}{
  search[men cold machine wash shirts black]
}

\casefield{Observation $o_1$}{
  A search-result page with 50 products is returned.
  \textcolor{decisionred}{Candidate \texttt{B09QQP3356} is visible at \$10.99}.
}

\end{tabularx}

{\scriptsize\textbf{Task State $z_1$:}}

\begin{lstlisting}[style=casestate]
stage: result
current_product: none
selected_options: []
search_history: [men cold machine wash shirts black]
visited_products: []
rejected_products: []
visited_navigation: []
recent_actions: [search[men cold machine wash shirts black]]
recent_failures: []
@@missing_constraints: [product, color=black, fit_type=youth, size=4t]@@
\end{lstlisting}

\end{tcolorbox}

\begin{tcolorbox}[casestep,title={Step 1}]

\begin{tabularx}{\linewidth}{
  @{}>{\raggedright\arraybackslash\bfseries}p{0.20\linewidth}
  >{\raggedright\arraybackslash}X@{}
}

\casefield{Active Skill Node}{
  \texttt{s1}
}

\casefield{Subgraph Decision}{
  Continue the active skill node through
  $
  \texttt{s1}
  \xrightarrow{\mathtt{decomposes\_to}}
  \texttt{o1}
  $.
}

\casefield{Selected Target}{
  \texttt{o1}: click(asin=B09QQP3356)
}

\casefield{Grounded Action}{
  click[b09qqp3356]
}

\casefield{Observation $o_2$}{
  The product-detail page is opened.
  Black is available, but
  \textcolor{decisionred}{size 4t and fit type youth are absent.}
}

\end{tabularx}

{\scriptsize\textbf{Task State $z_2$:}}

\begin{lstlisting}[style=casestate]
stage: detail
current_product: B09QQP3356
selected_options: []
search_history: [men cold machine wash shirts black]
visited_products: [B09QQP3356]
rejected_products: []
visited_navigation: []
recent_actions: [search[men cold machine wash shirts black], click[b09qqp3356]]
@@recent_failures: [candidate_missing_size=4t, candidate_missing_fit_type=youth]@@
@@missing_constraints: [fit_type=youth, size=4t]@@
\end{lstlisting}

\end{tcolorbox}

\begin{tcolorbox}[casestep,title={Step 2}]

\begin{tabularx}{\linewidth}{
  @{}>{\raggedright\arraybackslash\bfseries}p{0.20\linewidth}
  >{\raggedright\arraybackslash}X@{}
}

\casefield{Active Skill Node}{
    \texttt{s3}
}

\casefield{Subgraph Decision}{
  Execution stuck, switch to a recovery skill node through:
  $
  \texttt{s3}
  \xrightarrow{\mathtt{recovers\_with}}
  \texttt{s1}
  $.
  Then select
  $
  \texttt{s3}
  \xrightarrow{\mathtt{decomposes\_to}}
  \texttt{o1}
  $.
}

\casefield{Selected Target}{
  \texttt{o1}: click(target=back to search)
}

\casefield{Grounded Action}{
  click[back to search]
}

\casefield{Observation $o_3$}{
  The search page is displayed again.
}

\end{tabularx}

{\scriptsize\textbf{Task State $z_3$:}}

\begin{lstlisting}[style=casestate]
stage: search
current_product: none
selected_options: []
search_history: [men cold machine wash shirts black]
visited_products: [B09QQP3356]
rejected_products: [B09QQP3356 -> {missing_size=4t, missing_fit_type=youth}]
@@visited_navigation: [back to search]@@
recent_actions: [search[men cold machine wash shirts black], click[b09qqp3356], click[back to search]]
@@recent_failures: [candidate_missing_size=4t, candidate_missing_fit_type=youth]@@
@@missing_constraints: [product, fit_type=youth, size=4t]@@
\end{lstlisting}

\end{tcolorbox}

\begin{tcolorbox}[casestep,title={Step 3}]

\begin{tabularx}{\linewidth}{
  @{}>{\raggedright\arraybackslash\bfseries}p{0.20\linewidth}
  >{\raggedright\arraybackslash}X@{}
}

\casefield{Active Skill Node}{
  \texttt{s3}
}

\casefield{Subgraph Decision}{
  Continue the active skill node through
  $
  \texttt{s3}
  \xrightarrow{\mathtt{decomposes\_to}}
  \texttt{o0}
  $.
}

\casefield{Selected Target}{
  \texttt{o0}: search(query=men shirts black 4t)
}

\casefield{Grounded Action}{
  search[men shirts black 4t]
}

\casefield{Observation $o_4$}{
  A refined result page is returned. \textcolor{decisionred}{Candidate \texttt{B09P39QN2W} appears at \$16.99}.
}

\end{tabularx}

{\scriptsize\textbf{Task State $z_4$:}}

\begin{lstlisting}[style=casestate]
stage: result
current_product: none
search_history: [men cold machine wash shirts black, men shirts black 4t]
visited_products: [B09QQP3356]
rejected_products: [B09QQP3356 -> {missing_size=4t, missing_fit_type=youth}]
visited_navigation: [back to search]
recent_actions: [search[men cold machine wash shirts black], click[b09qqp3356], click[back to search], search[men shirts black 4t]]
recent_failures: []
@@missing_constraints: [product, color=black, fit_type=youth, size=4t]@@
\end{lstlisting}

\end{tcolorbox}

\begin{tcolorbox}[casestep,title={Step 4}]

\begin{tabularx}{\linewidth}{
  @{}>{\raggedright\arraybackslash\bfseries}p{0.20\linewidth}
  >{\raggedright\arraybackslash}X@{}
}

\casefield{Active Skill Node}{
  \texttt{s9}
}

\casefield{Subgraph Decision}{
  Select a successor skill node through
  $
  \texttt{s3}
  \xrightarrow{\mathtt{can\_follow}}
  \texttt{s9}
  $.
  Then select
  $
  \texttt{s9}
  \xrightarrow{\mathtt{decomposes\_to}}
  \texttt{o6}
  $.
}

\casefield{Selected Target}{
  \texttt{o6}: click\_product(asin=B09P39QN2W)
}

\casefield{Grounded Action}{
  click[b09p39qn2w]
}

\casefield{Observation $o_5$}{
  The new product provides \textcolor{decisionred}{youth, black, and 4t options, costs
  \$16.99}, and exposes \textcolor{decisionred}{Buy Now}.
}

\end{tabularx}

{\scriptsize\textbf{Task State $z_5$:}}

\begin{lstlisting}[style=casestate]
stage: detail
@@current_product: B09P39QN2W@@
selected_options: []
search_history: [men cold machine wash shirts black, men shirts black 4t]
visited_products: [B09QQP3356, B09P39QN2W]
rejected_products: [B09QQP3356 -> {missing_size=4t, missing_fit_type=youth}]
visited_navigation: [back to search]
recent_actions: [click[b09qqp3356], click[back to search], search[men shirts black 4t], click[b09p39qn2w]]
recent_failures: []
@@missing_constraints: [color=black, fit_type=youth, size=4t]@@
\end{lstlisting}

\end{tcolorbox}

\begin{tcolorbox}[casestep,title={Step 5}]

\begin{tabularx}{\linewidth}{
  @{}>{\raggedright\arraybackslash\bfseries}p{0.20\linewidth}
  >{\raggedright\arraybackslash}X@{}
}

\casefield{Active Skill Node}{
  \texttt{s6}
}

\casefield{Subgraph Decision}{
  Select a successor skill node through
  $
  \texttt{s9}
  \xleftrightarrow{\mathtt{compatible\_with}}
  \texttt{s6}
  $.
  Then select
  $
  \texttt{s6}
  \xrightarrow{\mathtt{decomposes\_to}}
  \texttt{o2}
  $.
}

\casefield{Selected Target}{
  \texttt{o2}: select\_color(value=black)
}

\casefield{Grounded Action}{
  click[black]
}

\casefield{Observation $o_6$}{
  \textcolor{decisionred}{Color black is available and selected}.
}

\end{tabularx}

{\scriptsize\textbf{Task State $z_6$:}}

\begin{lstlisting}[style=casestate]
stage: detail
current_product: B09P39QN2W
@@selected_options: [black]@@
search_history: [men cold machine wash shirts black, men shirts black 4t]
visited_products: [B09QQP3356, B09P39QN2W]
rejected_products: [B09QQP3356 -> {missing_size=4t, missing_fit_type=youth}]
visited_navigation: [back to search]
recent_actions: [click[back to search], search[men shirts black 4t], click[b09p39qn2w], click[black]]
recent_failures: []
@@missing_constraints: [fit_type=youth, size=4t]@@
\end{lstlisting}

\end{tcolorbox}

\begin{tcolorbox}[casestep,title={Step 6}]

\begin{tabularx}{\linewidth}{
  @{}>{\raggedright\arraybackslash\bfseries}p{0.20\linewidth}
  >{\raggedright\arraybackslash}X@{}
}

\casefield{Active Skill Node}{
  \texttt{s6}
}

\casefield{Subgraph Decision}{
  Continue the active skill node through
  $
  \texttt{s6}
  \xrightarrow{\mathtt{decomposes\_to}}
  \texttt{o3}
  $.
}

\casefield{Selected Target}{
  \texttt{o3}: select\_size(value=4t)
}

\casefield{Grounded Action}{
  click[4t]
}

\casefield{Observation $o_7$}{
  \textcolor{decisionred}{Size 4t is selected}.
}

\end{tabularx}

{\scriptsize\textbf{Task State $z_7$:}}

\begin{lstlisting}[style=casestate]
stage: detail
current_product: B09P39QN2W
@@selected_options: [black, 4t]@@
search_history: [men cold machine wash shirts black, men shirts black 4t]
visited_products: [B09QQP3356, B09P39QN2W]
rejected_products: [B09QQP3356 -> {missing_size=4t, missing_fit_type=youth}]
visited_navigation: [back to search]
recent_actions: [search[men shirts black 4t], click[b09p39qn2w], click[black], click[4t]]
recent_failures: []
@@missing_constraints: [fit_type=youth]@@
\end{lstlisting}

\end{tcolorbox}

\begin{tcolorbox}[casestep,title={Step 7}]

\begin{tabularx}{\linewidth}{
  @{}>{\raggedright\arraybackslash\bfseries}p{0.20\linewidth}
  >{\raggedright\arraybackslash}X@{}
}

\casefield{Active Skill Node}{
  \texttt{s6}
}

\casefield{Subgraph Decision}{
  The missing fit constraint, select the support AtomicOp node through
  $
  \texttt{o1}
  \xrightarrow{\mathtt{supports}}
  \texttt{s6}
  $.
}

\casefield{Selected Target}{
  \texttt{o1}: click(value=youth)
}

\casefield{Grounded Action}{
  click[youth]
}

\casefield{Observation $o_8$}{
  \textcolor{decisionred}{Fit type youth is selected. \textcolor{decisionred}{All required options are satisfied}, and Buy Now
  remains available.}
}

\end{tabularx}

{\scriptsize\textbf{Task State $z_8$:}}

\begin{lstlisting}[style=casestate]
stage: detail
current_product: B09P39QN2W
@@selected_options: [black, 4t, youth]@@
search_history: [men cold machine wash shirts black, men shirts black 4t]
visited_products: [B09QQP3356, B09P39QN2W]
rejected_products: [B09QQP3356 -> {missing_size=4t, missing_fit_type=youth}]
visited_navigation: [back to search]
recent_actions: [click[b09p39qn2w], click[black], click[4t], click[youth]]
recent_failures: []
@@missing_constraints: []@@
\end{lstlisting}

\end{tcolorbox}

\begin{tcolorbox}[casestep,title={Step 8}]

\begin{tabularx}{\linewidth}{
  @{}>{\raggedright\arraybackslash\bfseries}p{0.20\linewidth}
  >{\raggedright\arraybackslash}X@{}
}

\casefield{Active Skill Node}{
  \texttt{s6}
}

\casefield{Subgraph Decision}{
  All constraints are satisfied, and continue the active skill node through
  $
  \texttt{s6}
  \xrightarrow{\mathtt{decomposes\_to}}
  \texttt{o4}
  $.
}

\casefield{Selected Target}{
  \texttt{o4}: buy(asin=B09P39QN2W, size=4t, color=black, fit\_type=youth)
}

\casefield{Grounded Action}{
  click[buy now]
}

\casefield{Observation $o_9$}{
  The \textcolor{decisionred}{purchase succeeds} and the page reports a score of 100.0.
}

\end{tabularx}

{\scriptsize\textbf{Task State $z_9$:}}

\begin{lstlisting}[style=casestate]
@@stage: done@@
current_product: B09P39QN2W
@@selected_options: [black, 4t, youth]@@
search_history: [men cold machine wash shirts black, men shirts black 4t]
visited_products: [B09QQP3356, B09P39QN2W]
rejected_products: [B09QQP3356 -> {missing_size=4t, missing_fit_type=youth}]
visited_navigation: [back to search]
recent_actions: [click[black], click[4t], click[youth], click[buy now]]
recent_failures: []
@@missing_constraints: []@@
\end{lstlisting}

\end{tcolorbox}
\end{document}